\PassOptionsToPackage{table}{xcolor}
\PassOptionsToPackage{hyperfootnotes=false}{hyperref}
\documentclass[11pt]{article}
\usepackage[final]{acl}
\usepackage{times}
\usepackage{latexsym}
\usepackage[T1]{fontenc}
\usepackage[utf8]{inputenc}
\usepackage{microtype}
\usepackage{inconsolata}
\usepackage{booktabs}
\usepackage{tabularx}
\usepackage{longtable}
\usepackage{cuted}
\usepackage{capt-of}
\let\appendixendstrip\endstrip
\renewcommand{\endstrip}{\appendixendstrip\null\par}
\usepackage{graphicx}
\usepackage{xcolor}
\usepackage{url}
\usepackage{multirow}
\usepackage{array}

\newcolumntype{Y}{>{\raggedright\arraybackslash}X}
\newcolumntype{Z}{>{\raggedright\arraybackslash}X}
\newcolumntype{C}[1]{>{\raggedright\arraybackslash}m{#1}}
\definecolor{TableLit}{HTML}{E9F2EC}
\definecolor{TableIdea}{HTML}{EEF4EF}
\definecolor{TableExec}{HTML}{EDF1F5}
\definecolor{TableE2E}{HTML}{F0EFF5}
\definecolor{TableWriting}{HTML}{F5F1E9}
\definecolor{TablePeer}{HTML}{EBF3F2}

\title{How Is Automated Research Evaluated?\\A Survey of Benchmarks and Evaluation Practices}
\author{
  Liulei Zhang\thanks{Equal contribution.} \quad
  Dejing Zhou\footnotemark[1] \quad
  Chuyue Huang \quad Guanhua Chen \\
  \bfseries Yutong Yao \quad Lidia S. Chao \quad Chi Man Vong \quad
  Derek F. Wong\thanks{Corresponding author.} \\
  NLP\textsuperscript{2}CT Lab, Department of Artificial Intelligence \\
  University of Macau, Macau, China \\
  \small\texttt{nlp2ct.\{liulei,chuyue,guanhua,yutong\}@gmail.com} \\
  \small\texttt{\{mc55091,lidiasc,cmvong,derekfw\}@um.edu.mo}
}

\begin{document}
\maketitle

\begin{abstract}
Automated research systems support literature synthesis, ideation, experiments, writing, and peer review, but their evaluation is dispersed across tasks, benchmarks, and studies that are difficult to compare directly. We review this literature from the perspective of evaluation design and evidence, covering six targets: literature synthesis, research ideation, executable workflows, scholarly writing and communication, automatic peer review, and end-to-end research. We compare task construction, evidence sources, evaluators, and scoring procedures to explain the capabilities assessed by different designs. Our synthesis highlights three recurring lessons: output checks, process checks, and human studies provide complementary information; evaluator calibration is specific to the property being assessed; and resource budgets and attempt selection are integral to interpreting performance comparisons. We identify diagnostic evaluation designs and documented gaps in supporting evidence, and translate these comparisons into reporting and audit recommendations for specific evaluation settings. The survey helps readers navigate existing evaluations, select appropriate benchmarks, and design subsequent studies.
\end{abstract}

\section{Introduction}
Research agents search papers, formulate proposals, run experiments, write, and review manuscripts. Correct computations, useful literature comparisons, and supported criticisms require different observations. Interpreting their benchmark scores therefore requires understanding the task, available evidence, and scoring procedure.

Existing surveys and perspectives organize automated research around the scientific method, workflow stages, autonomy, applications, and system architectures \citep{zhang2025exploring,luo2025llm4sr,chen2025ai4research,zheng2025automation,wei2025ai,xie2025far,ren2025towards}. Our review brings together the evaluation literature within this landscape: benchmarks, datasets, execution environments, human studies, metrics, and audit protocols. We compare their experimental designs as well as reported scores.

Across six evaluation targets, we distinguish questions requiring different evidence, such as a review's usefulness to its author versus its detection of a known error. Figure~\ref{fig:claim-strength} maps representative works by target and claims supported by their core evaluation designs.

The survey makes three contributions. First, it \emph{organizes and reviews evaluation studies}, explaining their tasks and assessment procedures. Second, it \emph{compares evaluation designs and synthesizes findings}, connecting evidence, scoring, and calibration to assessed capabilities and support gaps. Third, it develops \emph{setting-specific recommendations} for benchmark selection, result interpretation, and evaluation design. The appendix records protocols and reporting details.

\section{Scope and Review Method}
\label{sec:scope}
We review evaluation of automated research workflows, their artifacts, and human interaction through explicit tasks or inspectable scoring procedures. Standalone retrieval, single-paper question answering, generic coding, and autonomous wet-lab systems are outside the main coverage; selected adjacent tasks contribute distinctive designs to core comparisons.

\paragraph{Search and selection.}
We searched Google Scholar, arXiv, Semantic Scholar, and the ACL Anthology iteratively alongside writing, with backward and forward citation tracing. The cutoff was May 10, 2026; descriptions of some studies incorporate subsequent revisions.

Queries combined target and evaluation terms (Appendix~\ref{sec:protocol-records}). This selective review places representative comparisons in the main text and additional designs and details in the appendix. Cross-target studies are classified by their primary evaluation question.

\paragraph{Analysis fields.}
We compare the \emph{target} (capability, artifact, or stage), \emph{evidence source} (judgment materials and observations), \emph{evaluator} (person, model, or procedure), \emph{metric} (scoring and aggregation), and \emph{validity boundary} (applicability conditions and identified claim--evidence gaps). Evidence-source analysis considers both provenance and support for output judgments. Related RAG studies examine evidence-based output calibration \citep{DBLP:conf/acl/0006YC0W25} and multi-granularity retrieval for verification \citep{DBLP:journals/corr/abs-2605-15019}. We use these fields to check evidence and scoring against authors' stated claims, distinguish observations from proxies, and examine calibration, identifying conclusions supported under the reported conditions.

Five claim-strength levels organize interpretation. \emph{L1: Surface plausibility} concerns artifact style or appearance. \emph{L2: Task correctness under specified evidence} concerns success under defined inputs and checks. \emph{L3: Process-grounded competence} requires criterion-checked process evidence explaining execution or decisions, beyond stored logs. \emph{L4: Human- or utility-calibrated performance} relates results to defined populations' judgments or outcomes, distinguishing preference, factual assessment, and practical benefit. \emph{L5: Scientific contribution} concerns advances supported by domain-appropriate novelty, methods, results, interpretation, and validation.

Cross-stage interpretation separates local results, verified overall completion, and stage interactions; process records explain interactions but are not universally required for reporting observable final outcomes.

\begin{figure*}[t]
\centering
\includegraphics[width=\textwidth]{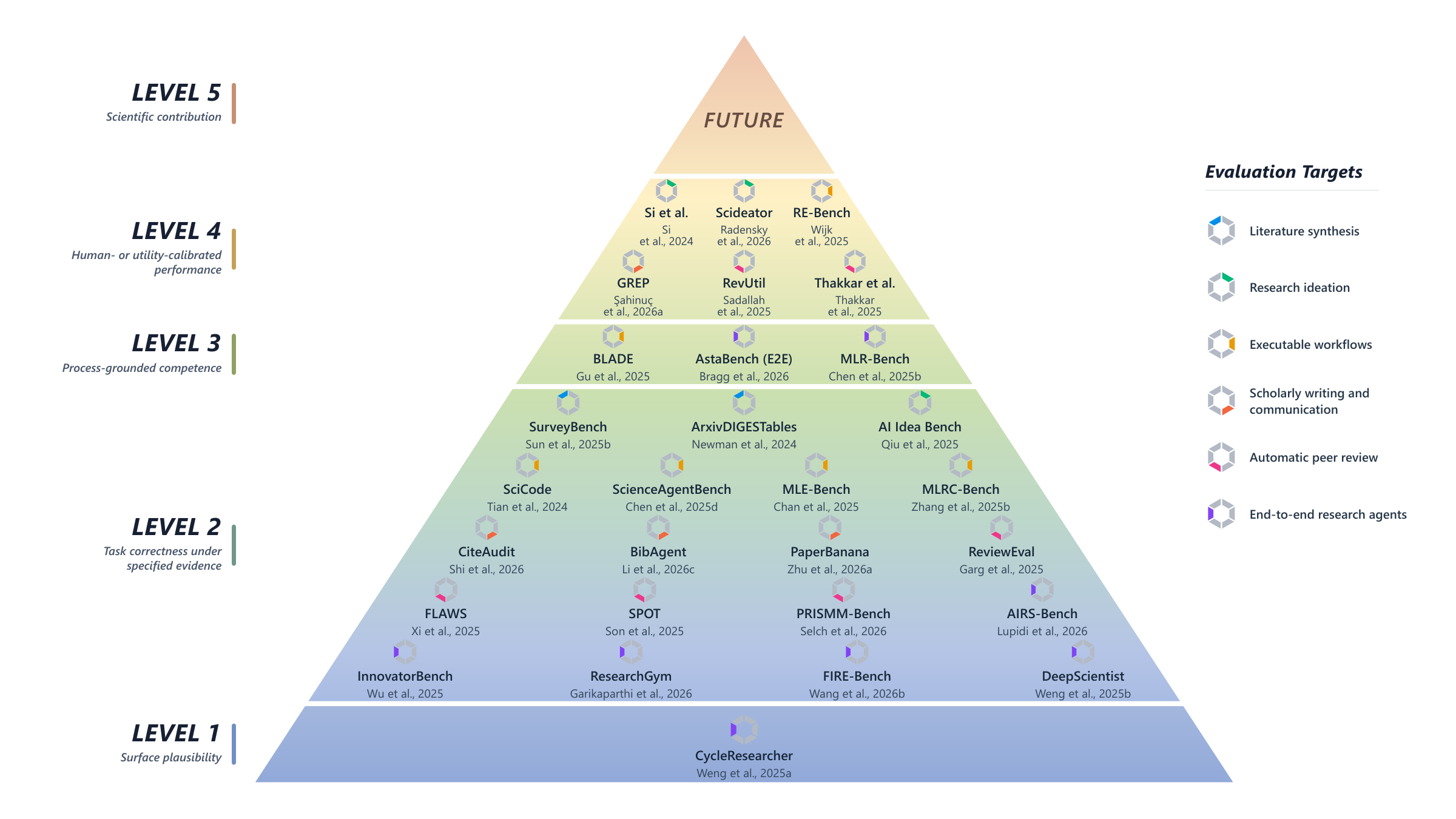}
\caption{Representative works by claim strength. Highlighted hexagon segments indicate primary evaluation targets. AstaBench (E2E) denotes only the suite's end-to-end component. Level 5 indicates a direction for further evaluation development.}
\label{fig:claim-strength}
\end{figure*}

\paragraph{Classifying representative works.}
Figure~\ref{fig:claim-strength} classifies core evaluation designs and main experiments; supplementary process analyses, human ratings, and scorer audits do not independently determine placement. Works without a single-level summary of equally central designs are omitted. Table~\ref{tab:compact-mapping} and Appendix~\ref{sec:protocol-records} detail selected classifications and protocols.

Most selected works assess task correctness (L2). BLADE, AstaBench (E2E), and MLR-Bench check analysis decisions or research steps (L3); RE-Bench and Scideator center expert-performance or user-judgment comparisons (L4).

\section{Target-Specific Evaluation}
\label{sec:targets}
We compare representative task designs and synthesize findings and open questions, interpreting specific protocols and reported results.

\subsection{Scientific Literature Synthesis Evaluation}

Literature synthesis turns multiple papers into an organized account of a research area or an answer to a research question. Evaluation must therefore specify who selects the sources, what the system produces, and how that output is checked. Long surveys, comparison tables, taxonomies, and question-driven answers expose different parts of this task; they are complementary evaluation settings, not successive stages of a single benchmark.

\paragraph{Survey generation and content assessment.}
AutoSurvey retrieves papers, builds an outline, and writes and revises individual sections \citep{wang2024autosurvey}. Its evaluation separates citation support from coverage, structure, and relevance. Importantly, its citation recall measures support among sentences already containing citations, rather than recovery of a human bibliography. SurGE instead uses 205 human surveys and a shared retrieval pool to compare reference coverage, citation accuracy, structure, and content \citep{su2025surge}. Its expert ranking study calibrates structure and content scores; it does not validate the separate citation judgments, which use titles and abstracts. Thus, similarly named citation metrics can have different denominators and evidence access. SurveyGen makes another important condition explicit: its 120-survey experiment distinguishes topic-only generation, automatic retrieval, and writing with human-provided references and outlines \citep{bao2025surveygen}. The last condition tests writing with supplied organization, not autonomous source selection.

\paragraph{Structured synthesis.}
Comparison tables require choosing useful dimensions and filling them with paper-specific evidence. ArxivDIGESTables evaluates these operations separately: schema recall measures recovery of human comparison columns, while cell evaluation supplies the human schema to isolate information extraction \citep{newman2024arxivdigestables}. Human checks cover 50 tables for column alignment and 30 for cell matching. Additional context mainly improves partial cell matches; generated columns absent from the reference are not significantly less useful in the reported study. Reference recovery and usefulness are therefore separately observable. TaxoBench similarly distinguishes finding expert-cited papers from organizing a supplied paper set into a hierarchy \citep{zhang2026can}. Its reported structural controls show that semantic path scores respond strongly to tree depth and category wording, so raw alignment differences require interpretation alongside these controls.

\paragraph{Question-driven synthesis and information use.}
OpenScholar evaluates citation-bearing scientific answers using expert questions, rubrics, and preferences; its expert comparison covers 108 questions and 16 evaluators \citep{asai2024openscholar}. ResearchQA scales a related task by mining questions and scoring requirements from surveys across 75 fields, with expert validation in eight fields \citep{yifei2025researchqa}. These designs differ in how information needs are obtained and how widely human calibration extends. SurveyBench instead asks whether an already generated survey supplies the information needed for general and topic-specific quizzes \citep{sun2025surveybench}. Across 20 topics, an LLM answers from retrieved survey passages, and unsupported answers receive zero. DeepResearch scores 4.42/5 on content but 1.97/10 on topic-specific questions; LLM$\times$MapReduce-V2 scores 4.20 and 3.19, respectively. This reversal shows why content ratings and supported answerability should both be inspected.

DeepSurvey-Bench adds a rubric for information, scholarly communication, and research guidance \citep{zhang2026deepsurveybench}. It reports human calibration on 40 generated surveys and a separate evidence audit sampled from high-scoring surveys. That audit describes evidence quality within the selected group. More generally, \citet{hwang2026deep} find that system-level agreement with experts can conceal weaker instance-level agreement in scientific long-answer evaluation.

\paragraph{Synthesis.}
The most informative designs separate source selection, organization, and use. Shared retrieval pools and supplied-outline conditions help identify which part of the task a comparison tests. Separate schema and cell evaluations similarly make an unsuccessful table easier to diagnose. ArxivDIGESTables' useful non-reference columns motivate assessing alternative structures alongside reference alignment; SurveyBench's differing content and answerability rankings show why covered information must also be usable for answering questions.

Evaluation should match intended use: coverage and organization for field overviews, supported answerability for questions, and citation checks with explicit evidence access and denominators for source-grounded claims. Calibration must follow these distinctions. Expert agreement on content, structure, or preference does not validate separate citation or individual-answer scoring. Reporting these dimensions separately would reveal which component accounts for a system's strengths or failures.

\subsection{Research Ideation and Proposal Evaluation}

Research ideas are evaluated before their proposed experiments have produced results. Novelty asks how an idea differs from existing work; feasibility concerns whether its steps can be carried out; value concerns the expected importance of the contribution. Existing protocols approach these questions through expert judgments, comparison with published ideas, retrieved prior work, and evidence-bearing proposal reports.

\paragraph{Expert judgments and published references.}
\citet{si2024can} compare human ideas, automatically selected AI ideas, and human-reranked AI ideas in seven prompting-based NLP topics. Their study involves 104 distinct researchers and 49 proposals per condition, with blinded ratings after style normalization. At the idea level, AI proposals receive higher novelty scores than human proposals, while feasibility differences are not significant. This is a comparison of the specified generation and selection processes: AI uses extensive sampling and retrieval, whereas each human contributes one idea. AI Idea Bench asks a different question \citep{qiu2025ai}. It supplies inspiration papers and a topic for 3,495 target papers, then evaluates generated motivations and experimental plans through target-paper matching and pairwise comparisons. Topic alignment, recovery of a published idea, and expert-perceived novelty consequently have different reference objects, even when all are used to evaluate ideation.

\paragraph{Novelty relative to prior work.}
Literature-Grounded Novelty Assessment makes the comparison set explicit \citep{shahid2025literaturegrounded}. It retrieves candidate papers, reranks them by purpose, mechanism, evaluation, and application, and judges novelty against the top ten using expert examples. On 32 held-out ideas, its reported accuracy is 0.81 and agreement with expert labels is $\kappa=0.59$. A separate 58-case study tests only known non-novel ideas and should not be read as overall classification accuracy. Navigating Ideation Space instead separates problem, method, and findings representations, providing component-level links to prior work \citep{shen2026navigating}. Its novelty score correlates with 93 expert-rated ideas, but the reported Spearman correlation of 0.384 is close to GPT-5 Mini's 0.385. Scideator brings related facet-based reasoning into an interactive tool \citep{radensky2026humanllm}. Its user study finds higher creativity-support ratings from 22 researchers, measuring tool experience rather than expert novelty labels.

\paragraph{Feasibility and proposal feedback.}
AI Idea Bench also derives overall and per-step feasibility proxies from the normalized, time-weighted citation impact of papers linked to decomposed idea steps. The authors interpret the per-step score as methodological soundness. However, citation impact does not test whether a retrieved method applies to the present step or whether steps are compatible; independent calibration against step-feasibility judgments is not reported. These scores directly measure the influence of associated literature.

ScholarEval and InnoEval seek richer assessments by producing evidence-bearing feedback \citep{moussa2025scholareval,qiao2026innoeval}. ScholarEval separates soundness and contribution analysis, evaluates coverage of 1,076 expert-checked feedback requirements, and studies report usefulness with 18 researchers. InnoEval combines papers, web resources, and code with multiple simulated reviewer perspectives; it evaluates historical decision labels, rankings, and expert dimension scores.

\paragraph{Synthesis.}
Ideation evaluation is most interpretable when novelty, feasibility, and feedback usefulness remain distinct. Blinding and style normalization control presentation; links to prior literature expose the basis of novelty judgments; expert-feedback requirements make coverage observable. These designs support interpretation without requiring agreement with a single published idea.

The unresolved feasibility question is whether influential supporting literature establishes that the current plan is implementable. AI Idea Bench's citation-derived proxies provide evidence about the associated papers. Its per-step methodological-soundness interpretation lacks calibration against method applicability and step compatibility. Such judgments would test the proposed connection between its citation-derived proxies and feasibility. Practitioners could combine literature-grounded novelty assessment with separately calibrated feasibility judgments and feedback-usefulness studies. Expert comparisons should also report generation and selection budgets, which determine the proposals reaching evaluators.

\subsection{Executable Research Workflow and Scientific Coding Evaluation}

Executable workflow evaluation covers scientific calculations, data analysis, and machine-learning research engineering. Systems receive tasks and computational resources and return code, predictions, or analysis outputs. Running these artifacts enables direct checks, but the checks differ: successful execution, correct scientific outputs, defensible analysis decisions, and improved experimental performance are distinct targets.

\paragraph{Scientific code and analysis.}
SciCode uses scientist-curated problems with numerical and domain-specific tests \citep{tian2024scicode}. Its 65-problem test set contains 288 subproblems, allowing local implementations and their integration to be compared. Claude 3.5 Sonnet solves 26.0\% of subproblems but 4.6\% of main problems in the standard setting; supplied scientific background raises these to 35.4\% and 12.3\%. These conditions help diagnose knowledge and integration requirements. Because experts provide the decomposition, the results test implementation along that workflow rather than the system's ability to invent the decomposition.

ScienceAgentBench moves to self-contained Python programs for 102 data-driven scientific tasks \citep{chen2025scienceagentbench}. It separates valid execution from task-specific success, using numerical checks, prediction thresholds, and, for some visual outputs, model scoring. For Claude's self-debugging configuration without expert knowledge, valid execution reaches 92.2\%, but task success is 32.4\%, under best-of-three selection. BLADE instead evaluates choices among defensible analyses: conceptual variables, data transformations, and statistical models are matched to expert reference decisions \citep{gu2024blade}. Its finding that executable analyses can still cover few reference decision combinations illustrates what a run/no-run check misses. DiscoveryBench scores final hypotheses by matching their conditions, variables, and relationships \citep{majumder2024discoverybench}. Its no-data baseline is diagnostic: Llama-3 scores 11.5 without data versus 12.1 with CodeGen. This identifies a contribution from question and metadata information; the matching score alone does not isolate data verification.

\paragraph{Machine-learning engineering and research methods.}
MLE-bench packages 75 Kaggle competitions as offline tasks \citep{chan2025mlebench}. Given a description and data, agents train models and submit predictions to a local hidden-test grader. Historical private-leaderboard thresholds translate task scores into medal attainment. Under AIDE, a 24-hour budget, and one A10 GPU, o1-preview attains a medal threshold in 16.9\% of attempts; allowing eight attempts raises the at-least-one-success rate to 34.1\%. Runtime and hardware specify conditions, while contamination checks audit the evidence. Because many tests are resplit from original training data, historical medals are calibrated references: the authors check example solutions and acknowledge that reconstructed and original competitions are not exactly equivalent.

RE-Bench supplies a complementary human comparison: 61 experts make 71 eight-hour attempts in seven research-engineering environments \citep{wijk2024rebench}. Relative performance changes with the budget. The reported 32-hour human result selects the best of four independent eight-hour attempts, rather than observing one continuous 32-hour session. This design makes time and attempt allocation explicit parts of the comparison.

MLRC-Bench connects proposed methods to implemented results in seven research competitions \citep{zhang2025mlrcbench}. Agents modify a protected repository, select a snapshot on development performance, and undergo hidden-test evaluation. Scores normalize improvement relative to baseline and human solutions and can be negative. Its auxiliary o1 novelty ratings have Spearman correlation $-0.06$ with measured effectiveness. Under this protocol, perceived innovativeness is a poor predictor of competition performance; effectiveness is not itself a novelty label, so the correlation does not establish that novelty judgments are inaccurate.

\paragraph{Synthesis.}
Executable evaluation gains diagnostic value by separating running code, obtaining correct results, and making appropriate analysis decisions. SciCode's local and integrated tests and ScienceAgentBench's execution and success measures locate different failures. BLADE checks analysis decisions against expert alternatives. DiscoveryBench's no-data control shows why performance must also be examined against information available without execution. Figure~\ref{fig:claim-strength} distinguishes task checks (L2), direct analysis-decision evaluation (L3), and RE-Bench's controlled expert comparison (L4).

For research engineering, hidden-test scores provide an endpoint, while attempt selection, feedback, and compute define its conditions. MLE-bench and RE-Bench show why best-of-several outcomes and single-attempt success answer different operational questions. Comparisons should preserve both task success and resource cost. Decision-level checks add evidence about how an analysis was conducted. Their role depends on whether the intended use is output verification, workflow diagnosis, or performance comparison under a shared budget.

\subsection{Academic Writing and Scholarly Communication Evaluation}

Scientific writing evaluation asks whether a system expresses research accurately, uses sources appropriately, and communicates through a particular medium. These questions require different observations: reconstructing a citation paragraph, checking a reference, and reading a poster are distinct tasks. Their shared concern is how information survives selection, attribution, and presentation.

\paragraph{Source-supported writing.}
CiteBench standardizes citation-sentence and paragraph generation from supplied papers and citing context \citep{funkquist2022citebench}. It combines reference similarity with citation-intent and discourse distributions, and human ratings of readability and consistency. In its 150-output human study, average consistency remained below 3/5 across models, despite readable outputs. Reference reconstruction also depends on input sufficiency: the authors document targets containing facts absent from the permitted input, although the affected proportion is unquantified. Thus, some reconstruction errors cannot be attributed solely to the generator. At introduction level, SciIG separately evaluates reference-set recovery, content, and narrative quality from supplied metadata \citep{garg2025lets}. The results show that its keyword-overlap measures and human factuality judgments have weak alignment, illustrating why metric names alone do not establish what they measure.

Task manipulation helps explain these differences. Across two models, six prompts, and six input configurations, \citet{sahinuc2024systematic} vary abstracts, citation intent, and example citations. Their exploratory 30-example human study finds that automatic measures provide only partial agreement with atomic-fact coverage. Reference-derived intent is an explicit experimental input, allowing the study to examine how task definition changes output. GREP instead evaluates related-work sections against hard constraints, such as source support, and soft preferences, such as citation emphasis and positioning \citep{sahinuc2025expert}. Its protocol includes 16 NLP/CV experts and criterion-specific comparisons. An initial 20\% faithfulness pass rate for o3-mini means that every citation passed in 20\% of sections, not that only 20\% of citations were faithful. Reported citation-format parsing failures also enter this score, so it does not isolate semantic reasoning failures.

\paragraph{Reference and citation verification.}
Bibliographic validity and contextual support need different evidence. BibAgent retrieves source passages to diagnose miscitation; its paywall setting instead uses downstream citing papers as indirect evidence and can abstain \citep{li2026bibagent}. MisciteBench contains 6,350 expert-checked synthetic miscitations, all erroneous. Its Acc-pass@3 requires a correct validity label and an accepted explanation in at least one of three attempts; it therefore does not estimate false alarms on naturally correct citations. CiteAudit tests existence and metadata consistency on separate generated and real-world collections containing valid and invalid references \citep{shi2026citeauditb}. Its reported precision falls from .947 to .749 while recall remains .971 and .951, respectively. Separate confusion matrices expose differences that a single accuracy or synthetic-data result would conceal.

\paragraph{Visual communication.}
Paper2Poster's PaperQuiz asks six vision-language models to answer source-paper questions using only a poster \citep{pang2025paper2poster}. This measures information accessibility alongside visual and textual quality; raw answer accuracy is separate from its length-adjusted score. Human calibration involved one PhD participant and five papers, limiting evidence for a general human-reader proxy. PaperBanana instead compares method diagrams on faithfulness, conciseness, readability, and aesthetics, with a hierarchical overall judgment \citep{zhu2026paperbanana}. Separate 50-example studies calibrate the judge and collect blind preferences. Its overall improvement coexists with faithfulness below the reference baseline.

\paragraph{Synthesis.}
Writing evaluation becomes actionable when criteria identify revision needs. GREP separates unsupported content from unwanted emphasis or presentation choices. Citation-generation findings show that reference-based errors must be interpreted against the writer's inputs; separating parsing from semantic verification would likewise distinguish formatting failures from support failures. These distinctions help turn an overall writing score into a diagnosis that can guide revision.

Readability, faithfulness, and information access can diverge, as SciIG's weak metric--human alignment and PaperBanana's differing overall and faithfulness results illustrate. Reference verification additionally requires detecting errors without false alarms on valid references. Assessment should therefore combine criteria suited to the artifact and its readers: source support and author constraints for prose, and fidelity and information accessibility for visual summaries. Human calibration should test the particular reader judgment represented by each automated measure.

\subsection{Automatic Peer Review Evaluation}

A useful scientific review connects criticism to manuscript evidence and gives its recipient something interpretable or actionable. Evaluation accordingly examines review content, the validity of specific criticisms, and effects within the review process. The central design choice is which property the available reference or observation can check.

\paragraph{Review content and evaluator calibration.}
ReviewEval decomposes quality into alignment with human reviews, factual accuracy, analytical depth, constructiveness, and guideline adherence \citep{garg2025revieweval}. Its factuality pipeline converts comments into questions, retrieves manuscript passages, generates a rebuttal, and checks support. Five experts also compare anonymous reviews on 30 papers, but their four preference dimensions do not supply independent labels for that question-level verifier. Consequently, high automatic support rates have different evidential backing from the reported human preferences. Peerispect makes verification explicit through supported, partially supported, contradicted, and undetermined labels \citep{ghorbanpour2026peerispect}. Its constructed manuscript-claim set contains only supported claims; the separate real-review set supplies the multi-class test. GEM addresses another evaluator question: how much information a review shares with independent peer references when no unique gold review exists \citep{xu2025benchmarkinga}. Teacher-score correlations and ICLR review-degradation tests are separate experiments. GEM is sensitive to information removal; unlike GPT-4o LMExaminer, its scores do not significantly increase after meaningless expansion in those tests. However, it does not outperform the LLM judge on every calibration measure.

\paragraph{Criticism and error localization.}
Natural weaknesses and controlled errors offer complementary reference construction strategies. CLAIMCHECK links expert-selected weaknesses to manuscript claims and separates claim association, weakness labeling/editing, and claim verification \citep{ou2025claimchecka}. Labeling/editing supplies the weakness; verification does not. The performance difference between these settings makes the input distinction essential. FLAWS constructs claim-invalidating edits and evaluates ranked error locations in 713 paper--error pairs \citep{xi2025flaws}. Its top-k score uses candidate depth within one output, with length restrictions on proposed spans. Human inspection validates 57 of 58 sampled errors, while a separate comparison checks automatic detection judgments. This provides evidence about both the constructed problems and their scorer.

SPOT instead contains 91 author-acknowledged errors in 83 historical manuscripts \citep{son2025when}. Models receive interleaved text and images; their descriptions are matched to known errors, and repeated runs support pass@K estimates. However, expert case analysis identifies a genuine unit error outside the annotations. Since unmatched findings count as false positives, matching precision cannot directly estimate the factual reliability of every criticism. PRISMM-Bench further isolates cross-modal inconsistencies through identification, remedy selection, and visual matching \citep{selch2026prismmbench}. Focused, page, and document contexts distinguish local recognition from broader inspection; no-context and structured-option controls reveal how answer wording can support guessing.

\paragraph{Usefulness and workflow effects.}
RevUtil evaluates author-oriented actionability, grounding/specificity, verifiability, and helpfulness using 1,430 human-annotated review points \citep{sadallah2025good}. Its scorer sees the comment, not the paper: grounding means pointing to manuscript content, while verifiability concerns supplied reasons. A randomized ICLR deployment observes a different endpoint by providing feedback to reviewers after submission \citep{thakkar2025can}. Of 22,467 reviews assigned to feedback, 18,946 received it after gating. Review edits and discussion behavior are measured separately from blind preferences on 100 selected high-uptake revision pairs. The reported 89\% preference for revisions concerns that subset.

\paragraph{Synthesis.}
Peer-review evaluation combines three views: complete-review quality, criticism correctness, and feedback effects. Multidimensional assessment, controlled or historical errors, and utility or deployment studies provide complementary evidence about content, detection, and use. Controlled errors make detection and localization testable, while historical errors retain problems encountered in real manuscripts. Human-annotated utility criteria and randomized deployment connect content to recipient judgments and workflow behavior. No single reference set supplies all three.

Calibration must match the scored property. A preference study assesses perceived review quality; an automated factuality checker needs independent judgments of manuscript support. ReviewEval illustrates this distinction, while FLAWS separately validates constructed errors and detection judgments. SPOT's valid finding outside its annotations shows why unmatched criticism needs adjudication before matching precision can represent factual reliability. Workflow studies add practical effects, but conclusions remain tied to feedback assignment, receipt, and the preference-study subset. Combining these checks would evaluate useful, well-grounded reviews while keeping content assessment, factual verification, and workflow effects separately interpretable.

\subsection{End-to-End Automatic Scientific Research System Evaluation}

End-to-end research evaluation follows a connected sequence of research decisions and outputs. The sequence depends on the task: some systems start with code templates, others with a research question or dataset. A multi-task suite may include such sequences alongside isolated tasks. Understanding the starting resources and final success criterion is therefore necessary before comparing overall scores.

\paragraph{Systems and evaluation environments.}
The AI Scientist connects ideation, experiments, figures, writing, and automated review within lightweight ML templates \citep{lu2024aiscientist}. Its counts of proposed ideas, completed experiments, and compiled papers expose different failure points. DeepScientist studies sustained exploration from reproduced baselines, combining task-native improvements, implementation records, and expert assessments \citep{weng2025deepscientist}. Its reported results include human verification and substantial compute. These examples motivate reusable protocols while retaining each system's starting points and verification arrangements.

AstaBench provides 11 benchmarks spanning literature, code, analysis, and end-to-end discovery, with tool and cost metadata \citep{bragg2026astabench}. The suite's aggregate is not a single connected research run. Its E2E-Bench and Hard variant evaluate reports, code, and artifacts against required research steps. A step passes only when evidence does not conflict and at least one source supports completion. The original study's Table 21 reports that 16\% of rubric-criterion assessments appeared satisfied in the paper but were contradicted by code or artifacts. Separately, a human spot check judged 92\% of 50 development-set rubric scores to be correct. Mean step scores and all-required-step completion are reported separately; the non-executing Faker control receives partial step credit but completes no whole task. Cross-artifact checks thus identify inconsistencies that report-only evaluation would miss.

\paragraph{Research artifacts and executable outcomes.}
MLR-Bench uses workshop-derived open research questions and phase-specific rubrics \citep{chen2025mlrbench}. Its 201 tasks support idea and proposal evaluation, whereas experiments, writing, and end-to-end runs mainly use ten selected tasks. MLR-Judge averages two model assessments and inspects execution records as well as papers. The authors report unverified synthetic or placeholder-data results in eight of ten Claude Code tasks, a concrete reason to inspect execution. Ten experts provide calibration, but non-significant differences between human--model and human--human score discrepancies do not establish statistical equivalence.

ResearchGym supplies runnable repositories with source methods removed, retained baselines, and task-native graders \citep{garikaparthi2026researchgym}. Under the initial 12-hour, single-A100 and \$10 API budget, its main agent improves on the strongest visible baseline in one of 15 runs; mean subtask completion is 26.5\%. Reporting completion, improvement, and best@3 separately distinguishes reliable execution from occasional gains. AIRS-Bench supplies data and evaluation scripts without initial solution code, accepting prediction files for hidden-label scoring \citep{lupidi2026airsbench}. Its 24-hour, single-H200 repeated runs offer a contrasting starting condition, so results should be compared through both task design and resource settings.

\paragraph{Rediscovery and stage diagnosis.}
FIRE-Bench asks agents to recover established empirical findings while withholding the original implementation and conclusions \citep{wang2026firebench}. For example, a question about information position can target the finding that relevant evidence is used better at the beginning or end of context than in the middle. The study runs 30 core and ten cross-domain problems, not the entire 100-paper source pool. Atomic-claim precision, recall, and F1 compare conclusions with reference findings; trajectory inspection attributes errors to planning, implementation, execution, and conclusion formation. Its added human check validates claim extraction, not the entire subsequent entailment scorer.

\paragraph{Synthesis.}
End-to-end evaluation needs explicit starting resources, required stages, and success criteria. Supplied templates, runnable baselines, and data-only starting points impose different research demands. Reporting stage and whole-task completion identifies where connected runs stop; hidden-test performance or reference-finding recovery measures their outcomes. Comparisons should retain these distinctions rather than assume equivalent starting points.

Cross-artifact verification makes completion judgments more diagnostic. AstaBench's report--artifact disagreements and MLR-Bench's placeholder-data cases show that code and experimental records can change the assessment of an apparently completed report. Process inspection tests whether claimed steps occurred and helps attribute failures; its findings should remain distinct from primary outcome scores, as with rediscovery matching and trajectory analysis. Evaluations should report starting assistance, verified completion, task-native outcomes, and resource use together, retaining successful runs and failures rather than only an aggregate artifact score.

\section{Toward Credible Evaluation Infrastructure}
\label{sec:agenda}
The comparisons motivate task-dependent reporting (Table~\ref{tab:reporting-template}): input-sufficiency issues in CiteBench, the need for feasibility-proxy calibration in AI Idea Bench, problem and scorer checks in FLAWS, and report--artifact verification in AstaBench.

\paragraph{Comparability and calibration.}
Resource conditions and repeated attempts shape results: MLE-bench varies attempt budgets, and RE-Bench compares agents and experts at stated time budgets. Long-software-task research provides background on task duration and success measurement \citep{kwa2025longtasks}. Evaluators also need checks: general LLM-judge studies document position and verbosity biases and examine possible self-enhancement bias \citep{zheng2023mtbench}. Their occurrence and impact require verification in each scientific setting.

\paragraph{Reporting recommendations.}
Report the task and intended claim, inputs and data version, evaluator and scoring, sample and aggregation unit, and applicability conditions. Combine applicable modules in Table~\ref{tab:reporting-template}, such as error construction and matching for localization, or participants and outcomes for feedback utility. Public artifacts support inspection and reuse.

\section{Conclusion}
Across six targets, this survey compares evaluation designs, synthesizes findings and evidence gaps, and develops setting-specific reporting recommendations. These comparisons help readers select benchmarks, interpret results, and design subsequent evaluations.

\section*{Limitations}
This review is selective and emphasizes automated research workflows in AI/ML with inspectable evaluation evidence. It does not exhaust the broader AI-for-Science or autonomous wet-lab literature. Coverage is also uneven across disciplines because many available benchmarks concern computer science and machine learning. We synthesize reported studies and documented protocols rather than independently reproducing their experiments. The five claim distinctions guide interpretation; they are not a validated measurement scale or an exhaustive taxonomy of scientific value.

\section*{Acknowledgements}
This work was supported in part by the Science and Technology Development Fund of Macau SAR (Grant Nos. FDCT/0007/2024/AKP, EF2024-00185-FST), the UM and UMDF (Grant Nos. MYRG-GRG2024-00165-FST-UMDF, MYRG-GRG2025-00236-FST, MYRG-GRG2026-00018-FIC, MYRG-CRG2026-00007-FIC), the Dr. Stanley Ho Medical Development Foundation (Grant No. SHMDF-AI/2026/001), and the National Natural Science Foundation of China (Grant No. 62266013).

\bibliography{references-1}

@inproceedings{DBLP:conf/acl/0006YC0W25,
  author       = {Guanhua Chen and
                  Yutong Yao and
                  Lidia S. Chao and
                  Xuebo Liu and
                  Derek F. Wong},
  editor       = {Wanxiang Che and
                  Joyce Nabende and
                  Ekaterina Shutova and
                  Mohammad Taher Pilehvar},
  title        = {{SGIC:} {A} Self-Guided Iterative Calibration Framework for {RAG}},
  booktitle    = {Proceedings of the 63rd Annual Meeting of the Association for Computational
                  Linguistics (Volume 1: Long Papers), {ACL} 2025, Vienna, Austria,
                  July 27 - August 1, 2025},
  pages        = {28357--28370},
  publisher    = {Association for Computational Linguistics},
  year         = {2025},
  url          = {https://aclanthology.org/2025.acl-long.1376/},
  bibsource    = {dblp computer science bibliography, https://dblp.org}
}

@article{DBLP:journals/corr/abs-2605-15019,
  author       = {Guanhua Chen and
                  Chuyue Huang and
                  Yutong Yao and
                  Shudong Liu and
                  Xueqing Song and
                  Lidia S. Chao and
                  Derek F. Wong},
  title        = {From Scenes to Elements: Multi-Granularity Evidence Retrieval for
                  Verifiable Multimodal {RAG}},
  journal      = {CoRR},
  volume       = {abs/2605.15019},
  year         = {2026},
  url          = {https://doi.org/10.48550/arXiv.2605.15019},
  doi          = {10.48550/ARXIV.2605.15019},
  eprinttype   = {arXiv},
  eprint       = {2605.15019},
  bibsource    = {dblp computer science bibliography, https://dblp.org}
}

@misc{chang2025sridbench,
      title={SridBench: Benchmark of Scientific Research Illustration Drawing of Image Generation Model}, 
      author={Yifan Chang and Yukang Feng and Jianwen Sun and Jiaxin Ai and Chuanhao Li and S. Kevin Zhou and Kaipeng Zhang},
      year={2025},
      eprint={2505.22126},
      archivePrefix={arXiv},
      primaryClass={cs.CV},
      url={https://arxiv.org/abs/2505.22126}, 
}

@inproceedings{chen2021scixgen,
  author       = {Hong Chen and
                  Hiroya Takamura and
                  Hideki Nakayama},
  editor       = {Marie{-}Francine Moens and
                  Xuanjing Huang and
                  Lucia Specia and
                  Scott Wen{-}tau Yih},
  title        = {SciXGen: {A} Scientific Paper Dataset for Context-Aware Text Generation},
  booktitle    = {Findings of the Association for Computational Linguistics: {EMNLP}
                  2021, Virtual Event / Punta Cana, Dominican Republic, 16-20 November,
                  2021},
  series       = {Findings of {ACL}},
  volume       = {{EMNLP} 2021},
  pages        = {1483--1492},
  publisher    = {Association for Computational Linguistics},
  year         = {2021},
  url          = {https://doi.org/10.18653/v1/2021.findings-emnlp.128},
  doi          = {10.18653/V1/2021.FINDINGS-EMNLP.128},
  bibsource    = {dblp computer science bibliography, https://dblp.org}
}

@misc{docekal2024oarelatedwork,
      title={OARelatedWork: A Large-Scale Dataset of Related Work Sections with Full-texts from Open Access Sources}, 
      author={Martin Docekal and Martin Fajcik and Pavel Smrz},
      year={2026},
      eprint={2405.01930},
      archivePrefix={arXiv},
      primaryClass={cs.CL},
      url={https://arxiv.org/abs/2405.01930}, 
}

@misc{funkquist2022citebench,
      title={CiteBench: A benchmark for Scientific Citation Text Generation}, 
      author={Martin Funkquist and Ilia Kuznetsov and Yufang Hou and Iryna Gurevych},
      year={2023},
      eprint={2212.09577},
      archivePrefix={arXiv},
      primaryClass={cs.CL},
      url={https://arxiv.org/abs/2212.09577}, 
}

@misc{garg2025lets,
      title={Towards AI-Assisted Research Writing: Benchmarking LLMs for AI/ML Introduction Generation}, 
      author={Krishna Garg and Firoz Shaik and Sambaran Bandyopadhyay and Cornelia Caragea},
      year={2026},
      eprint={2508.14273},
      archivePrefix={arXiv},
      primaryClass={cs.CL},
      url={https://arxiv.org/abs/2508.14273}, 
}

@misc{huang2026scifig,
      title={SciFig: Towards Automating Editable Figure Generation for Scientific Papers}, 
      author={Siyuan Huang and Yifan Zhou and Yutong Gao and Zi Yin and Juyang Bai and Xinxin Liu and Rama Chellappa and Chun Pong Lau and Cheng Peng and Sayan Nag and Shraman Pramanick},
      year={2026},
      eprint={2601.04390},
      archivePrefix={arXiv},
      primaryClass={cs.AI},
      url={https://arxiv.org/abs/2601.04390}, 
}

@inproceedings{nishimura2024structured,
    title = "Toward Structured Related Work Generation with Novelty Statements",
    author = "Nishimura, Kazuya  and
      Saito, Kuniaki  and
      Hirasawa, Tosho  and
      Ushiku, Yoshitaka",
    editor = "Ghosal, Tirthankar  and
      Singh, Amanpreet  and
      Waard, Anita  and
      Mayr, Philipp  and
      Naik, Aakanksha  and
      Weller, Orion  and
      Lee, Yoonjoo  and
      Shen, Shannon  and
      Qin, Yanxia",
    booktitle = "Proceedings of the Fourth Workshop on Scholarly Document Processing (SDP 2024)",
    month = aug,
    year = "2024",
    address = "Bangkok, Thailand",
    publisher = "Association for Computational Linguistics",
    url = "https://aclanthology.org/2024.sdp-1.5/",
    pages = "38--57"
}

@misc{pang2025paper2poster,
      title={Paper2Poster: Towards Multimodal Poster Automation from Scientific Papers}, 
      author={Wei Pang and Kevin Qinghong Lin and Xiangru Jian and Xi He and Philip Torr},
      year={2025},
      eprint={2505.21497},
      archivePrefix={arXiv},
      primaryClass={cs.CV},
      url={https://arxiv.org/abs/2505.21497}, 
}

@inproceedings{sahinuc2024systematic,
  author       = {Furkan Sahinu{\c{c}} and
                  Ilia Kuznetsov and
                  Yufang Hou and
                  Iryna Gurevych},
  editor       = {Lun{-}Wei Ku and
                  Andre Martins and
                  Vivek Srikumar},
  title        = {Systematic Task Exploration with LLMs: {A} Study in Citation Text
                  Generation},
  booktitle    = {Proceedings of the 62nd Annual Meeting of the Association for Computational
                  Linguistics (Volume 1: Long Papers), {ACL} 2024, Bangkok, Thailand,
                  August 11-16, 2024},
  pages        = {4832--4855},
  publisher    = {Association for Computational Linguistics},
  year         = {2024},
  url          = {https://doi.org/10.18653/v1/2024.acl-long.265},
  doi          = {10.18653/V1/2024.ACL-LONG.265},
  bibsource    = {dblp computer science bibliography, https://dblp.org}
}

@misc{sahinuc2025expert,
      title={Expert Preference-based Evaluation of Automated Related Work Generation}, 
      author={Furkan Şahinuç and Subhabrata Dutta and Iryna Gurevych},
      year={2026},
      eprint={2508.07955},
      archivePrefix={arXiv},
      primaryClass={cs.CL},
      url={https://arxiv.org/abs/2508.07955}, 
}

@misc{sahinuc2026reward,
      title={Reward Modeling for Scientific Writing Evaluation}, 
      author={Furkan Şahinuç and Subhabrata Dutta and Iryna Gurevych},
      year={2026},
      eprint={2601.11374},
      archivePrefix={arXiv},
      primaryClass={cs.CL},
      url={https://arxiv.org/abs/2601.11374}, 
}

@misc{sun2025p2p,
      title={P2P: Automated Paper-to-Poster Generation and Fine-Grained Benchmark}, 
      author={Tao Sun and Enhao Pan and Zhengkai Yang and Kaixin Sui and Jiajun Shi and Xianfu Cheng and Tongliang Li and Wenhao Huang and Ge Zhang and Jian Yang and Zhoujun Li},
      year={2025},
      eprint={2505.17104},
      archivePrefix={arXiv},
      primaryClass={cs.CL},
      url={https://arxiv.org/abs/2505.17104}, 
}

@inproceedings{weng2025cycleresearcher,
  author       = {Yixuan Weng and
                  Minjun Zhu and
                  Guangsheng Bao and
                  Hongbo Zhang and
                  Jindong Wang and
                  Yue Zhang and
                  Linyi Yang},
  title        = {CycleResearcher: Improving Automated Research via Automated Review},
  booktitle    = {The Thirteenth International Conference on Learning Representations,
                  {ICLR} 2025, Singapore, April 24-28, 2025},
  publisher    = {OpenReview.net},
  year         = {2025},
  url          = {https://openreview.net/forum?id=bjcsVLoHYs},
  bibsource    = {dblp computer science bibliography, https://dblp.org}
}

@misc{zeng2025slidetailor,
      title={SlideTailor: Personalized Presentation Slide Generation for Scientific Papers}, 
      author={Wenzheng Zeng and Mingyu Ouyang and Langyuan Cui and Hwee Tou Ng},
      year={2025},
      eprint={2512.20292},
      archivePrefix={arXiv},
      primaryClass={cs.CL},
      url={https://arxiv.org/abs/2512.20292}, 
}

@misc{zhu2026autofigure,
      title={AutoFigure: Generating and Refining Publication-Ready Scientific Illustrations}, 
      author={Minjun Zhu and Zhen Lin and Yixuan Weng and Panzhong Lu and Qiujie Xie and Yifan Wei and Sifan Liu and Qiyao Sun and Yue Zhang},
      year={2026},
      eprint={2602.03828},
      archivePrefix={arXiv},
      primaryClass={cs.AI},
      url={https://arxiv.org/abs/2602.03828}, 
}

@misc{zhu2026paperbanana,
      title={PaperBanana: Automating Academic Illustration for AI Scientists}, 
      author={Dawei Zhu and Rui Meng and Yale Song and Xiyu Wei and Sujian Li and Tomas Pfister and Jinsung Yoon},
      year={2026},
      eprint={2601.23265},
      archivePrefix={arXiv},
      primaryClass={cs.CL},
      url={https://arxiv.org/abs/2601.23265}, 
}

@misc{chan2025mlebench,
      title={MLE-bench: Evaluating Machine Learning Agents on Machine Learning Engineering}, 
      author={Jun Shern Chan and Neil Chowdhury and Oliver Jaffe and James Aung and Dane Sherburn and Evan Mays and Giulio Starace and Kevin Liu and Leon Maksin and Tejal Patwardhan and Lilian Weng and Aleksander Mądry},
      year={2025},
      eprint={2410.07095},
      archivePrefix={arXiv},
      primaryClass={cs.CL},
      url={https://arxiv.org/abs/2410.07095}, 
}

@misc{chen2025scienceagentbench,
      title={ScienceAgentBench: Toward Rigorous Assessment of Language Agents for Data-Driven Scientific Discovery}, 
      author={Ziru Chen and Shijie Chen and Yuting Ning and Qianheng Zhang and Boshi Wang and Botao Yu and Yifei Li and Zeyi Liao and Chen Wei and Zitong Lu and Vishal Dey and Mingyi Xue and Frazier N. Baker and Benjamin Burns and Daniel Adu-Ampratwum and Xuhui Huang and Xia Ning and Song Gao and Yu Su and Huan Sun},
      year={2025},
      eprint={2410.05080},
      archivePrefix={arXiv},
      primaryClass={cs.CL},
      url={https://arxiv.org/abs/2410.05080}, 
}

@misc{feng2025grapheval,
      title={GraphEval: A Lightweight Graph-Based LLM Framework for Idea Evaluation}, 
      author={Tao Feng and Yihang Sun and Jiaxuan You},
      year={2025},
      eprint={2503.12600},
      archivePrefix={arXiv},
      primaryClass={cs.LG},
      url={https://arxiv.org/abs/2503.12600}, 
}

@misc{gu2024blade,
      title={BLADE: Benchmarking Language Model Agents for Data-Driven Science}, 
      author={Ken Gu and Ruoxi Shang and Ruien Jiang and Keying Kuang and Richard-John Lin and Donghe Lyu and Yue Mao and Youran Pan and Teng Wu and Jiaqian Yu and Yikun Zhang and Tianmai M. Zhang and Lanyi Zhu and Mike A. Merrill and Jeffrey Heer and Tim Althoff},
      year={2025},
      eprint={2408.09667},
      archivePrefix={arXiv},
      primaryClass={cs.CL},
      url={https://arxiv.org/abs/2408.09667}, 
}

@misc{guo2024ideabench,
      title={IdeaBench: Benchmarking Large Language Models for Research Idea Generation}, 
      author={Sikun Guo and Amir Hassan Shariatmadari and Guangzhi Xiong and Albert Huang and Eric Xie and Stefan Bekiranov and Aidong Zhang},
      year={2024},
      eprint={2411.02429},
      archivePrefix={arXiv},
      primaryClass={cs.CL},
      url={https://arxiv.org/abs/2411.02429}, 
}

@misc{huang2023mlagentbench,
      title={MLAgentBench: Evaluating Language Agents on Machine Learning Experimentation}, 
      author={Qian Huang and Jian Vora and Percy Liang and Jure Leskovec},
      year={2024},
      eprint={2310.03302},
      archivePrefix={arXiv},
      primaryClass={cs.LG},
      url={https://arxiv.org/abs/2310.03302}, 
}

@misc{kon2025curie,
      title={Curie: Toward Rigorous and Automated Scientific Experimentation with AI Agents}, 
      author={Patrick Tser Jern Kon and Jiachen Liu and Qiuyi Ding and Yiming Qiu and Zhenning Yang and Yibo Huang and Jayanth Srinivasa and Myungjin Lee and Mosharaf Chowdhury and Ang Chen},
      year={2025},
      eprint={2502.16069},
      archivePrefix={arXiv},
      primaryClass={cs.AI},
      url={https://arxiv.org/abs/2502.16069}, 
}

@misc{kumar2025can,
      title={Can Large Language Models Unlock Novel Scientific Research Ideas?}, 
      author={Sandeep Kumar and Tirthankar Ghosal and Vinayak Goyal and Asif Ekbal},
      year={2025},
      eprint={2409.06185},
      archivePrefix={arXiv},
      primaryClass={cs.CL},
      url={https://arxiv.org/abs/2409.06185}, 
}

@misc{majumder2024discoverybench,
      title={DiscoveryBench: Towards Data-Driven Discovery with Large Language Models}, 
      author={Bodhisattwa Prasad Majumder and Harshit Surana and Dhruv Agarwal and Bhavana Dalvi Mishra and Abhijeetsingh Meena and Aryan Prakhar and Tirth Vora and Tushar Khot and Ashish Sabharwal and Peter Clark},
      year={2024},
      eprint={2407.01725},
      archivePrefix={arXiv},
      primaryClass={cs.CL},
      url={https://arxiv.org/abs/2407.01725}, 
}

@misc{nathani2025mlgym,
      title={MLGym: A New Framework and Benchmark for Advancing AI Research Agents}, 
      author={Deepak Nathani and Lovish Madaan and Nicholas Roberts and Nikolay Bashlykov and Ajay Menon and Vincent Moens and Amar Budhiraja and Despoina Magka and Vladislav Vorotilov and Gaurav Chaurasia and Dieuwke Hupkes and Ricardo Silveira Cabral and Tatiana Shavrina and Jakob Foerster and Yoram Bachrach and William Yang Wang and Roberta Raileanu},
      year={2025},
      eprint={2502.14499},
      archivePrefix={arXiv},
      primaryClass={cs.CL},
      url={https://arxiv.org/abs/2502.14499}, 
}

@misc{qiao2026innoeval,
      title={InnoEval: On Research Idea Evaluation as a Knowledge-Grounded, Multi-Perspective Reasoning Problem}, 
      author={Shuofei Qiao and Yunxiang Wei and Xuehai Wang and Bin Wu and Boyang Xue and Ningyu Zhang and Hossein A. Rahmani and Yanshan Wang and Qiang Zhang and Keyan Ding and Jeff Z. Pan and Huajun Chen and Emine Yilmaz},
      year={2026},
      eprint={2602.14367},
      archivePrefix={arXiv},
      primaryClass={cs.CL},
      url={https://arxiv.org/abs/2602.14367}, 
}

@misc{qiu2025ai,
      title={AI Idea Bench 2025: AI Research Idea Generation Benchmark}, 
      author={Yansheng Qiu and Haoquan Zhang and Zhaopan Xu and Ming Li and Diping Song and Zheng Wang and Kaipeng Zhang},
      year={2025},
      eprint={2504.14191},
      archivePrefix={arXiv},
      primaryClass={cs.AI},
      url={https://arxiv.org/abs/2504.14191}, 
}

@misc{radensky2026humanllm,
      title={Human-LLM Compound System for Scientific Ideation through Facet Recombination and Novelty Evaluation}, 
      author={Marissa Radensky and Simra Shahid and Raymond Fok and Pao Siangliulue and Tom Hope and Daniel S. Weld},
      year={2026},
      eprint={2409.14634},
      archivePrefix={arXiv},
      primaryClass={cs.HC},
      url={https://arxiv.org/abs/2409.14634}, 
}

@misc{ruan2026evaluating,
      title={Evaluating LLMs' Divergent Thinking Capabilities for Scientific Idea Generation with Minimal Context}, 
      author={Kai Ruan and Xuan Wang and Jixiang Hong and Peng Wang and Yang Liu and Hao Sun},
      year={2026},
      eprint={2412.17596},
      archivePrefix={arXiv},
      primaryClass={cs.CL},
      url={https://arxiv.org/abs/2412.17596}, 
}

@inproceedings{shahid2025literaturegrounded,
  author       = {Simra Shahid and
                  Marissa Radensky and
                  Raymond Fok and
                  Pao Siangliulue and
                  Daniel S. Weld and
                  Tom Hope},
  editor       = {Amanpreet Singh},
  title        = {Literature-Grounded Novelty Assessment of Scientific Ideas},
  booktitle    = {Proceedings of the Fifth Workshop on Scholarly Document Processing,
                  SDP@ACL 2025, Venice, Austria, July 31, 2025},
  pages        = {96--113},
  publisher    = {Association for Computational Linguistics},
  year         = {2025},
  url          = {https://aclanthology.org/2025.sdp-1.9/},
  bibsource    = {dblp computer science bibliography, https://dblp.org}
}

@misc{shen2026navigating,
      title={Navigating Ideation Space: Decomposed Conceptual Representations for Positioning Scientific Ideas}, 
      author={Yuexi Shen and Minqian Liu and Dawei Zhou and Lifu Huang},
      year={2026},
      eprint={2601.08901},
      archivePrefix={arXiv},
      primaryClass={cs.IR},
      url={https://arxiv.org/abs/2601.08901}, 
}

@misc{si2024can,
      title={Can LLMs Generate Novel Research Ideas? A Large-Scale Human Study with 100+ NLP Researchers}, 
      author={Chenglei Si and Diyi Yang and Tatsunori Hashimoto},
      year={2024},
      eprint={2409.04109},
      archivePrefix={arXiv},
      primaryClass={cs.CL},
      url={https://arxiv.org/abs/2409.04109}, 
}

@misc{sun2026dsaeval,
      title={DSAEval: Evaluating Data Science Agents on a Wide Range of Real-World Data Science Problems}, 
      author={Maojun Sun and Yifei Xie and Yue Wu and Ruijian Han and Binyan Jiang and Defeng Sun and Yancheng Yuan and Jian Huang},
      year={2026},
      eprint={2601.13591},
      archivePrefix={arXiv},
      primaryClass={cs.AI},
      url={https://arxiv.org/abs/2601.13591}, 
}

@misc{tian2024scicode,
      title={SciCode: A Research Coding Benchmark Curated by Scientists}, 
      author={Minyang Tian and Luyu Gao and Shizhuo Dylan Zhang and Xinan Chen and Cunwei Fan and Xuefei Guo and Roland Haas and Pan Ji and Kittithat Krongchon and Yao Li and Shengyan Liu and Di Luo and Yutao Ma and Hao Tong and Kha Trinh and Chenyu Tian and Zihan Wang and Bohao Wu and Yanyu Xiong and Shengzhu Yin and Minhui Zhu and Kilian Lieret and Yanxin Lu and Genglin Liu and Yufeng Du and Tianhua Tao and Ofir Press and Jamie Callan and Eliu Huerta and Hao Peng},
      year={2024},
      eprint={2407.13168},
      archivePrefix={arXiv},
      primaryClass={cs.AI},
      url={https://arxiv.org/abs/2407.13168}, 
}

@inproceedings{wang2024scimon,
  author       = {Qingyun Wang and
                  Doug Downey and
                  Heng Ji and
                  Tom Hope},
  editor       = {Lun{-}Wei Ku and
                  Andre Martins and
                  Vivek Srikumar},
  title        = {SciMON: Scientific Inspiration Machines Optimized for Novelty},
  booktitle    = {Proceedings of the 62nd Annual Meeting of the Association for Computational
                  Linguistics (Volume 1: Long Papers), {ACL} 2024, Bangkok, Thailand,
                  August 11-16, 2024},
  pages        = {279--299},
  publisher    = {Association for Computational Linguistics},
  year         = {2024},
  url          = {https://doi.org/10.18653/v1/2024.acl-long.18},
  doi          = {10.18653/V1/2024.ACL-LONG.18},
  bibsource    = {dblp computer science bibliography, https://dblp.org}
}

@misc{wijk2024rebench,
      title={RE-Bench: Evaluating frontier AI R\&D capabilities of language model agents against human experts}, 
      author={Hjalmar Wijk and Tao Lin and Joel Becker and Sami Jawhar and Neev Parikh and Thomas Broadley and Lawrence Chan and Michael Chen and Josh Clymer and Jai Dhyani and Elena Ericheva and Katharyn Garcia and Brian Goodrich and Nikola Jurkovic and Holden Karnofsky and Megan Kinniment and Aron Lajko and Seraphina Nix and Lucas Sato and William Saunders and Maksym Taran and Ben West and Elizabeth Barnes},
      year={2025},
      eprint={2411.15114},
      archivePrefix={arXiv},
      primaryClass={cs.LG},
      url={https://arxiv.org/abs/2411.15114}, 
}

@inproceedings{zhang2024benchmarking,
  author       = {Yuge Zhang and
                  Qiyang Jiang and
                  Xingyu Han and
                  Nan Chen and
                  Yuqing Yang and
                  Kan Ren},
  editor       = {Lun{-}Wei Ku and
                  Andre Martins and
                  Vivek Srikumar},
  title        = {Benchmarking Data Science Agents},
  booktitle    = {Proceedings of the 62nd Annual Meeting of the Association for Computational
                  Linguistics (Volume 1: Long Papers), {ACL} 2024, Bangkok, Thailand,
                  August 11-16, 2024},
  pages        = {5677--5700},
  publisher    = {Association for Computational Linguistics},
  year         = {2024},
  url          = {https://doi.org/10.18653/v1/2024.acl-long.308},
  doi          = {10.18653/V1/2024.ACL-LONG.308},
  bibsource    = {dblp computer science bibliography, https://dblp.org}
}

@misc{zhang2025mlrcbench,
      title={MLRC-Bench: Can Language Agents Solve Machine Learning Research Challenges?}, 
      author={Yunxiang Zhang and Muhammad Khalifa and Shitanshu Bhushan and Grant D Murphy and Lajanugen Logeswaran and Jaekyeom Kim and Moontae Lee and Honglak Lee and Lu Wang},
      year={2025},
      eprint={2504.09702},
      archivePrefix={arXiv},
      primaryClass={cs.AI},
      url={https://arxiv.org/abs/2504.09702}, 
}

@misc{zou2025fmlbench,
      title={FML-bench: Benchmarking Machine Learning Agents for Scientific Research}, 
      author={Qiran Zou and Hou Hei Lam and Wenhao Zhao and Yiming Tang and Tingting Chen and Samson Yu and Tianyi Zhang and Chang Liu and Xiangyang Ji and Dianbo Liu},
      year={2026},
      eprint={2510.10472},
      archivePrefix={arXiv},
      primaryClass={cs.CL},
      url={https://arxiv.org/abs/2510.10472}, 
}

@misc{dycke2026automatic,
      title={Automatic Reviewers Fail to Detect Faulty Reasoning in Research Papers: A New Counterfactual Evaluation Framework}, 
      author={Nils Dycke and Iryna Gurevych},
      year={2026},
      eprint={2508.21422},
      archivePrefix={arXiv},
      primaryClass={cs.CL},
      url={https://arxiv.org/abs/2508.21422}, 
}

@inproceedings{garg2025revieweval,
  author       = {Madhav Krishan Garg and
                  Tejash Prasad and
                  Tanmay Singhal and
                  Chhavi Kirtani and
                  Murari Mandal and
                  Dhruv Kumar},
  editor       = {Christos Christodoulopoulos and
                  Tanmoy Chakraborty and
                  Carolyn Rose and
                  Violet Peng},
  title        = {ReviewEval: An Evaluation Framework for AI-Generated Reviews},
  booktitle    = {Findings of the Association for Computational Linguistics: {EMNLP}
                  2025, Suzhou, China, November 4-9, 2025},
  pages        = {20542--20564},
  publisher    = {Association for Computational Linguistics},
  year         = {2025},
  url          = {https://doi.org/10.18653/v1/2025.findings-emnlp.1120},
  doi          = {10.18653/V1/2025.FINDINGS-EMNLP.1120},
  bibsource    = {dblp computer science bibliography, https://dblp.org}
}

@inproceedings{ghorbanpour2026peerispect,
   title={Peerispect: Claim Verification in Scientific Peer Reviews},
   url={http://dx.doi.org/10.1145/3805712.3808368},
   DOI={10.1145/3805712.3808368},
   booktitle={Proceedings of the 49th International ACM SIGIR Conference on Research and Development in Information Retrieval},
   publisher={ACM},
   author={Ghorbanpour, Ali and Sadeghian, Soroush and Daghighfarsoodeh, Alireza and Ebrahimi, Sajad and Arabzadeh, Negar and Hosseini, Seyed Mohammad and Bagheri, Ebrahim},
   year={2026},
   month=jul, pages={5161--5165} }

@misc{li2026rating,
      title={Beyond Rating: A Comprehensive Evaluation and Benchmark for AI Reviews}, 
      author={Bowen Li and Haochen Ma and Yuxin Wang and Jie Yang and Yining Zheng and Xinchi Chen and Xuanjing Huang and Xipeng Qiu},
      year={2026},
      eprint={2604.19502},
      archivePrefix={arXiv},
      primaryClass={cs.CL},
      url={https://arxiv.org/abs/2604.19502}, 
}

@misc{liang2023can,
      title={Can large language models provide useful feedback on research papers? A large-scale empirical analysis}, 
      author={Weixin Liang and Yuhui Zhang and Hancheng Cao and Binglu Wang and Daisy Ding and Xinyu Yang and Kailas Vodrahalli and Siyu He and Daniel Smith and Yian Yin and Daniel McFarland and James Zou},
      year={2023},
      eprint={2310.01783},
      archivePrefix={arXiv},
      primaryClass={cs.LG},
      url={https://arxiv.org/abs/2310.01783}, 
}

@inproceedings{ou2025claimchecka,
  author       = {Jiefu Ou and
                  William Gantt Walden and
                  Kate Sanders and
                  Zhengping Jiang and
                  Kaiser Sun and
                  Jeffrey Cheng and
                  William Jurayj and
                  Miriam Wanner and
                  Shaobo Liang and
                  Candice Morgan and
                  Seunghoon Han and
                  Weiqi Wang and
                  Chandler May and
                  Hannah Recknor and
                  Daniel Khashabi and
                  Benjamin Van Durme},
  editor       = {Christos Christodoulopoulos and
                  Tanmoy Chakraborty and
                  Carolyn Rose and
                  Violet Peng},
  title        = {{CLAIMCHECK:} How Grounded are {LLM} Critiques of Scientific Papers?},
  booktitle    = {Findings of the Association for Computational Linguistics: {EMNLP}
                  2025, Suzhou, China, November 4-9, 2025},
  pages        = {21712--21735},
  publisher    = {Association for Computational Linguistics},
  year         = {2025},
  url          = {https://doi.org/10.18653/v1/2025.findings-emnlp.1185},
  doi          = {10.18653/V1/2025.FINDINGS-EMNLP.1185},
  bibsource    = {dblp computer science bibliography, https://dblp.org}
}

@misc{ryu2025reviewscore,
      title={ReviewScore: Misinformed Peer Review Detection with Large Language Models}, 
      author={Hyun Ryu and Doohyuk Jang and Hyemin S. Lee and Joonhyun Jeong and Gyeongman Kim and Donghyeon Cho and Gyouk Chu and Minyeong Hwang and Hyeongwon Jang and Changhun Kim and Haechan Kim and Jina Kim and Joowon Kim and Yoonjeon Kim and Kwanhyung Lee and Chanjae Park and Heecheol Yun and Gregor Betz and Eunho Yang},
      year={2026},
      eprint={2509.21679},
      archivePrefix={arXiv},
      primaryClass={cs.CL},
      url={https://arxiv.org/abs/2509.21679}, 
}

@inproceedings{sadallah2025good,
  author       = {Abdelrahman Sadallah and
                  Tim Baumg{\"{a}}rtner and
                  Iryna Gurevych and
                  Ted Briscoe},
  editor       = {Christos Christodoulopoulos and
                  Tanmoy Chakraborty and
                  Carolyn Rose and
                  Violet Peng},
  title        = {The Good, the Bad and the Constructive: Automatically Measuring Peer
                  Review's Utility for Authors},
  booktitle    = {Proceedings of the 2025 Conference on Empirical Methods in Natural
                  Language Processing, {EMNLP} 2025, Suzhou, China, November 4-9, 2025},
  pages        = {28991--29021},
  publisher    = {Association for Computational Linguistics},
  year         = {2025},
  url          = {https://doi.org/10.18653/v1/2025.emnlp-main.1476},
  doi          = {10.18653/V1/2025.EMNLP-MAIN.1476},
  bibsource    = {dblp computer science bibliography, https://dblp.org}
}

@misc{selch2026prismmbench,
      title={PRISMM-Bench: A Benchmark of Peer-Review Grounded Multimodal Inconsistencies}, 
      author={Lukas Selch and Yufang Hou and M. Jehanzeb Mirza and Sivan Doveh and James Glass and Rogerio Feris and Wei Lin},
      year={2026},
      eprint={2510.16505},
      archivePrefix={arXiv},
      primaryClass={cs.CV},
      url={https://arxiv.org/abs/2510.16505}, 
}

@inproceedings{shin2025minda,
  author       = {Hyungyu Shin and
                  Jingyu Tang and
                  Yoonjoo Lee and
                  Nayoung Kim and
                  Hyunseung Lim and
                  Ji Yong Cho and
                  Hwajung Hong and
                  Moontae Lee and
                  Juho Kim},
  editor       = {Christos Christodoulopoulos and
                  Tanmoy Chakraborty and
                  Carolyn Rose and
                  Violet Peng},
  title        = {Mind the Blind Spots: {A} Focus-Level Evaluation Framework for {LLM}
                  Reviews},
  booktitle    = {Proceedings of the 2025 Conference on Empirical Methods in Natural
                  Language Processing, {EMNLP} 2025, Suzhou, China, November 4-9, 2025},
  pages        = {35630--35656},
  publisher    = {Association for Computational Linguistics},
  year         = {2025},
  url          = {https://doi.org/10.18653/v1/2025.emnlp-main.1805},
  doi          = {10.18653/V1/2025.EMNLP-MAIN.1805},
  bibsource    = {dblp computer science bibliography, https://dblp.org}
}

@misc{son2025when,
      title={When AI Co-Scientists Fail: SPOT-a Benchmark for Automated Verification of Scientific Research}, 
      author={Guijin Son and Jiwoo Hong and Honglu Fan and Heejeong Nam and Hyunwoo Ko and Seungwon Lim and Jinyeop Song and Jinha Choi and Gonçalo Paulo and Youngjae Yu and Stella Biderman},
      year={2025},
      eprint={2505.11855},
      archivePrefix={arXiv},
      primaryClass={cs.CL},
      url={https://arxiv.org/abs/2505.11855}, 
}

@misc{thakkar2025can,
      title={Can LLM feedback enhance review quality? A randomized study of 20K reviews at ICLR 2025}, 
      author={Nitya Thakkar and Mert Yuksekgonul and Jake Silberg and Animesh Garg and Nanyun Peng and Fei Sha and Rose Yu and Carl Vondrick and James Zou},
      year={2025},
      eprint={2504.09737},
      archivePrefix={arXiv},
      primaryClass={cs.AI},
      url={https://arxiv.org/abs/2504.09737}, 
}

@misc{xi2025flaws,
      title={FLAWS: A Benchmark for Error Identification and Localization in Scientific Papers}, 
      author={Sarina Xi and Vishisht Rao and Justin Payan and Nihar B. Shah},
      year={2025},
      eprint={2511.21843},
      archivePrefix={arXiv},
      primaryClass={cs.CL},
      url={https://arxiv.org/abs/2511.21843}, 
}

@misc{xu2025benchmarkinga,
      title={Benchmarking LLMs' Judgments with No Gold Standard}, 
      author={Shengwei Xu and Yuxuan Lu and Grant Schoenebeck and Yuqing Kong},
      year={2025},
      eprint={2411.07127},
      archivePrefix={arXiv},
      primaryClass={cs.CL},
      url={https://arxiv.org/abs/2411.07127}, 
}

@misc{xu2025can,
      title={Can LLMs Identify Critical Limitations within Scientific Research? A Systematic Evaluation on AI Research Papers}, 
      author={Zhijian Xu and Yilun Zhao and Manasi Patwardhan and Lovekesh Vig and Arman Cohan},
      year={2025},
      eprint={2507.02694},
      archivePrefix={arXiv},
      primaryClass={cs.CL},
      url={https://arxiv.org/abs/2507.02694}, 
}

@misc{zeng2023scientific,
      title={Scientific Opinion Summarization: Paper Meta-review Generation Dataset, Methods, and Evaluation}, 
      author={Qi Zeng and Mankeerat Sidhu and Ansel Blume and Hou Pong Chan and Lu Wang and Heng Ji},
      year={2024},
      eprint={2305.14647},
      archivePrefix={arXiv},
      primaryClass={cs.CL},
      url={https://arxiv.org/abs/2305.14647}, 
}

@misc{zhang2026detecting,
      title={Detecting Reference Errors in Scientific Literature with Large Language Models},
      author={Tianmai M. Zhang and Neil F. Abernethy},
      year={2024},
      eprint={2411.06101},
      archivePrefix={arXiv},
      primaryClass={cs.CL},
      url={https://arxiv.org/abs/2411.06101},
}

@inproceedings{zhou2024llm,
  author       = {Ruiyang Zhou and
                  Lu Chen and
                  Kai Yu},
  editor       = {Nicoletta Calzolari and
                  Min{-}Yen Kan and
                  V{\'{e}}ronique Hoste and
                  Alessandro Lenci and
                  Sakriani Sakti and
                  Nianwen Xue},
  title        = {Is {LLM} a Reliable Reviewer? {A} Comprehensive Evaluation of {LLM}
                  on Automatic Paper Reviewing Tasks},
  booktitle    = {Proceedings of the 2024 Joint International Conference on Computational
                  Linguistics, Language Resources and Evaluation, {LREC/COLING} 2024,
                  20-25 May, 2024, Torino, Italy},
  pages        = {9340--9351},
  publisher    = {{ELRA} and {ICCL}},
  year         = {2024},
  url          = {https://aclanthology.org/2024.lrec-main.816},
  bibsource    = {dblp computer science bibliography, https://dblp.org}
}

@misc{asai2024openscholar,
      title={OpenScholar: Synthesizing Scientific Literature with Retrieval-augmented LMs}, 
      author={Akari Asai and Jacqueline He and Rulin Shao and Weijia Shi and Amanpreet Singh and Joseph Chee Chang and Kyle Lo and Luca Soldaini and Sergey Feldman and Mike D'arcy and David Wadden and Matt Latzke and Minyang Tian and Pan Ji and Shengyan Liu and Hao Tong and Bohao Wu and Yanyu Xiong and Luke Zettlemoyer and Graham Neubig and Dan Weld and Doug Downey and Wen-tau Yih and Pang Wei Koh and Hannaneh Hajishirzi},
      year={2024},
      eprint={2411.14199},
      archivePrefix={arXiv},
      primaryClass={cs.CL},
      url={https://arxiv.org/abs/2411.14199}, 
}

@misc{bao2025surveygen,
      title={SurveyGen: Quality-Aware Scientific Survey Generation with Large Language Models}, 
      author={Tong Bao and Mir Tafseer Nayeem and Davood Rafiei and Chengzhi Zhang},
      year={2025},
      eprint={2508.17647},
      archivePrefix={arXiv},
      primaryClass={cs.CL},
      url={https://arxiv.org/abs/2508.17647}, 
}

@misc{guo2025sgsimeval,
      title={SGSimEval: A Comprehensive Multifaceted and Similarity-Enhanced Benchmark for Automatic Survey Generation Systems}, 
      author={Beichen Guo and Zhiyuan Wen and Yu Yang and Peng Gao and Ruosong Yang and Jiaxing Shen},
      year={2025},
      eprint={2508.11310},
      archivePrefix={arXiv},
      primaryClass={cs.CL},
      url={https://arxiv.org/abs/2508.11310}, 
}

@inproceedings{huang2025multidocument,
  author       = {Hui Huang and
                  Julien Velcin and
                  Yacine Kessaci},
  editor       = {Christos Christodoulopoulos and
                  Tanmoy Chakraborty and
                  Carolyn Rose and
                  Violet Peng},
  title        = {Towards Multi-Document Question Answering in Scientific Literature:
                  Pipeline, Dataset, and Evaluation},
  booktitle    = {Findings of the Association for Computational Linguistics: {EMNLP}
                  2025, Suzhou, China, November 4-9, 2025},
  pages        = {10867--10881},
  publisher    = {Association for Computational Linguistics},
  year         = {2025},
  url          = {https://doi.org/10.18653/v1/2025.findings-emnlp.576},
  doi          = {10.18653/V1/2025.FINDINGS-EMNLP.576},
  bibsource    = {dblp computer science bibliography, https://dblp.org}
}

@misc{hwang2026deep,
      title={Deep Research, Shallow Evaluation: A Case Study in Meta-Evaluation for Long-Form QA Benchmarks}, 
      author={Jena D. Hwang and Varsha Kishore and Amanpreet Singh and Dany Haddad and Aakanksha Naik and Malachi Hamada and Jonathan Bragg and Mike D'Arcy and Daniel S. Weld and Lucy Lu Wang and Doug Downey and Sergey Feldman},
      year={2026},
      eprint={2603.06942},
      archivePrefix={arXiv},
      primaryClass={cs.CL},
      url={https://arxiv.org/abs/2603.06942}, 
}

@misc{lahiri2025taxoalign,
      title={TaxoAlign: Scholarly Taxonomy Generation Using Language Models}, 
      author={Avishek Lahiri and Yufang Hou and Debarshi Kumar Sanyal},
      year={2025},
      eprint={2510.17263},
      archivePrefix={arXiv},
      primaryClass={cs.CL},
      url={https://arxiv.org/abs/2510.17263}, 
}

@misc{newman2024arxivdigestables,
      title={ArxivDIGESTables: Synthesizing Scientific Literature into Tables using Language Models}, 
      author={Benjamin Newman and Yoonjoo Lee and Aakanksha Naik and Pao Siangliulue and Raymond Fok and Juho Kim and Daniel S. Weld and Joseph Chee Chang and Kyle Lo},
      year={2024},
      eprint={2410.22360},
      archivePrefix={arXiv},
      primaryClass={cs.CL},
      url={https://arxiv.org/abs/2410.22360}, 
}

@misc{skarlinski2024language,
      title={Language agents achieve superhuman synthesis of scientific knowledge}, 
      author={Michael D. Skarlinski and Sam Cox and Jon M. Laurent and James D. Braza and Michaela Hinks and Michael J. Hammerling and Manvitha Ponnapati and Samuel G. Rodriques and Andrew D. White},
      year={2024},
      eprint={2409.13740},
      archivePrefix={arXiv},
      primaryClass={cs.CL},
      url={https://arxiv.org/abs/2409.13740}, 
}

@misc{su2025surge,
      title={SurGE: A Benchmark and Evaluation Framework for Scientific Survey Generation}, 
      author={Weihang Su and Anzhe Xie and Qingyao Ai and Jianming Long and Xuanyi Chen and Jiaxin Mao and Ziyi Ye and Yiqun Liu},
      year={2026},
      eprint={2508.15658},
      archivePrefix={arXiv},
      primaryClass={cs.CL},
      url={https://arxiv.org/abs/2508.15658}, 
}

@misc{sun2025surveybench,
      title={SurveyBench: Can LLM(-Agents) Write Academic Surveys that Align with Reader Needs?}, 
      author={Zhaojun Sun and Xuzhou Zhu and Xuanhe Zhou and Xin Tong and Shuo Wang and Jie Fu and Guoliang Li and Zhiyuan Liu and Fan Wu},
      year={2025},
      eprint={2510.03120},
      archivePrefix={arXiv},
      primaryClass={cs.CL},
      url={https://arxiv.org/abs/2510.03120}, 
}

@inproceedings{tangTangXueMei2025large,
  author       = {Xuemei Tang and
                  Xufeng Duan and
                  Zhenguang G. Cai},
  editor       = {Christos Christodoulopoulos and
                  Tanmoy Chakraborty and
                  Carolyn Rose and
                  Violet Peng},
  title        = {Large Language Models for Automated Literature Review: An Evaluation
                  of Reference Generation, Abstract Writing, and Review Composition},
  booktitle    = {Proceedings of the 2025 Conference on Empirical Methods in Natural
                  Language Processing, {EMNLP} 2025, Suzhou, China, November 4-9, 2025},
  pages        = {1602--1617},
  publisher    = {Association for Computational Linguistics},
  year         = {2025},
  url          = {https://doi.org/10.18653/v1/2025.emnlp-main.83},
  doi          = {10.18653/V1/2025.EMNLP-MAIN.83},
  bibsource    = {dblp computer science bibliography, https://dblp.org}
}

@misc{wang2024autosurvey,
      title={AutoSurvey: Large Language Models Can Automatically Write Surveys}, 
      author={Yidong Wang and Qi Guo and Wenjin Yao and Hongbo Zhang and Xin Zhang and Zhen Wu and Meishan Zhang and Xinyu Dai and Min Zhang and Qingsong Wen and Wei Ye and Shikun Zhang and Yue Zhang},
      year={2024},
      eprint={2406.10252},
      archivePrefix={arXiv},
      primaryClass={cs.IR},
      url={https://arxiv.org/abs/2406.10252}, 
}

@misc{wang2025arxiv2table,
      title={arXiv2Table: Toward Realistic Benchmarking and Evaluation for LLM-Based Literature-Review Table Generation}, 
      author={Weiqi Wang and Jiefu Ou and Yangqiu Song and Benjamin Van Durme and Daniel Khashabi},
      year={2026},
      eprint={2504.10284},
      archivePrefix={arXiv},
      primaryClass={cs.CL},
      url={https://arxiv.org/abs/2504.10284}, 
}

@misc{yan2025surveyforge,
      title={SurveyForge: On the Outline Heuristics, Memory-Driven Generation, and Multi-dimensional Evaluation for Automated Survey Writing}, 
      author={Xiangchao Yan and Shiyang Feng and Jiakang Yuan and Renqiu Xia and Bin Wang and Bo Zhang and Lei Bai},
      year={2025},
      eprint={2503.04629},
      archivePrefix={arXiv},
      primaryClass={cs.CL},
      url={https://arxiv.org/abs/2503.04629}, 
}

@misc{yifei2025researchqa,
      title={ResearchQA: Evaluating Scholarly Question Answering at Scale Across 75 Fields with Survey-Mined Questions and Rubrics}, 
      author={Li S. Yifei and Allen Chang and Chaitanya Malaviya and Mark Yatskar},
      year={2025},
      eprint={2509.00496},
      archivePrefix={arXiv},
      primaryClass={cs.CL},
      url={https://arxiv.org/abs/2509.00496}, 
}

@misc{zhang2026can,
      title={Can Deep Research Agents Retrieve and Organize? Evaluating the Synthesis Gap with Expert Taxonomies}, 
      author={Ming Zhang and Jiabao Zhuang and Wenqing Jing and Kexin Tan and Ziyu Kong and Jingyi Deng and Yujiong Shen and Yuhui Wang and Zhenghao Xiang and Qiyuan Peng and Yuhang Zhao and Ning Luo and Renzhe Zheng and Jiahui Lin and Mingqi Wu and Long Ma and Shihan Dou and Maxm Pan and Tao Gui and Qi Zhang and Xuanjing Huang},
      year={2026},
      eprint={2601.12369},
      archivePrefix={arXiv},
      primaryClass={cs.CL},
      url={https://arxiv.org/abs/2601.12369}, 
}

@misc{zhang2026deepsurveybench,
      title={DeepSurvey-Bench: Evaluating Academic Value of Automatically Generated Scientific Surveys}, 
      author={Guo-Biao Zhang and Xian-Ling Mao and Ding-Yuan Liu and Da-Yi Wu and Tian Lan and Huihui Li and Heyan Huang},
      year={2026},
      eprint={2601.15307},
      archivePrefix={arXiv},
      primaryClass={cs.AI},
      url={https://arxiv.org/abs/2601.15307}, 
}

@misc{zhao2025surveyeval,
      title={SurveyEval: Towards Comprehensive Evaluation of LLM-Generated Academic Surveys}, 
      author={Jiahao Zhao and Shuaixing Zhang and Nan Xu and Lei Wang},
      year={2025},
      eprint={2512.02763},
      archivePrefix={arXiv},
      primaryClass={cs.CL},
      url={https://arxiv.org/abs/2512.02763}, 
}

@Article{chelli2024hallucination,
author="Chelli, Mika{\"e}l
and Descamps, Jules
and Lavou{\'e}, Vincent
and Trojani, Christophe
and Azar, Michel
and Deckert, Marcel
and Raynier, Jean-Luc
and Clowez, Gilles
and Boileau, Pascal
and Ruetsch-Chelli, Caroline",
title="Hallucination Rates and Reference Accuracy of ChatGPT and Bard for Systematic Reviews: Comparative Analysis",
journal="J Med Internet Res",
year="2024",
month="May",
day="22",
volume="26",
pages="e53164",
issn="1438-8871",
doi="10.2196/53164",
url="https://www.jmir.org/2024/1/e53164"
}

@misc{li2026bibagent,
      title={BibAgent: An Agentic Framework for Traceable Miscitation Detection in Scientific Literature}, 
      author={Peiran Li and Fangzhou Lin and Shuo Xing and Xiang Zheng and Xi Hong and Siyuan Yang and Jiashuo Sun and Zhengzhong Tu and Chaoqun Ni},
      year={2026},
      eprint={2601.16993},
      archivePrefix={arXiv},
      primaryClass={cs.DL},
      url={https://arxiv.org/abs/2601.16993}, 
}

@misc{li2026source,
      title={Source or It Didn't Happen: A Multi-Agent Framework for Citation Hallucination Detection}, 
      author={Mingzhe Li and Zhiqiang Lin and Shiqing Ma},
      year={2026},
      eprint={2605.08583},
      archivePrefix={arXiv},
      primaryClass={cs.CL},
      url={https://arxiv.org/abs/2605.08583}, 
}

@misc{naser2026how,
      title={How LLMs Cite and Why It Matters: A Cross-Model Audit of Reference Fabrication in AI-Assisted Academic Writing and Methods to Detect Phantom Citations}, 
      author={MZ Naser},
      year={2026},
      eprint={2603.03299},
      archivePrefix={arXiv},
      primaryClass={cs.CL},
      url={https://arxiv.org/abs/2603.03299}, 
}

@misc{rao2026bibtex,
      title={BibTeX Citation Errors in Scientific Publishing Agents: Evaluation and Mitigation}, 
      author={Delip Rao and Chris Callison-Burch},
      year={2026},
      eprint={2604.03159},
      archivePrefix={arXiv},
      primaryClass={cs.DL},
      url={https://arxiv.org/abs/2604.03159}, 
}

@misc{sakai2026hallucitation,
      title={HalluCitation Matters: Revealing the Impact of Hallucinated References with 300 Hallucinated Papers in ACL Conferences}, 
      author={Yusuke Sakai and Hidetaka Kamigaito and Taro Watanabe},
      year={2026},
      eprint={2601.18724},
      archivePrefix={arXiv},
      primaryClass={cs.CL},
      url={https://arxiv.org/abs/2601.18724}, 
}

@misc{shi2026citeauditb,
      title={CiteAudit: You Cited It, But Did You Read It? A Benchmark for Verifying Scientific References in the LLM Era}, 
      author={Kaiwen Shi and Weixiang Sun and Zheyuan Zhang and Lichao Sun and Nitesh V. Chawla and Yanfang Ye},
      year={2026},
      eprint={2602.23452},
      archivePrefix={arXiv},
      primaryClass={cs.CL},
      url={https://arxiv.org/abs/2602.23452}, 
}

@article{walters2023fabrication,
	title = {Fabrication and errors in the bibliographic citations generated by {ChatGPT}},
	volume = {13},
	issn = {2045-2322},
	url = {https://doi.org/10.1038/s41598-023-41032-5},
	doi = {10.1038/s41598-023-41032-5},
	number = {1},
	journal = {Scientific Reports},
	author = {Walters, William H. and Wilder, Esther Isabelle},
	month = sep,
	year = {2023},
	pages = {14045},
}

@misc{xu2026ghostcite,
      title={GhostCite: A Large-Scale Analysis of Citation Validity in the Age of Large Language Models}, 
      author={Zuyao Xu and Yuqi Qiu and Lu Sun and Fasheng Miao and Fubin Wu and Xiang Li and Xinyi Wang and Haozhe Lu and Zhengze Zhang and Yuxin Hu and Jialu Li and Luo Jin and Feng Zhang and Rui Luo and Xinran Liu and Yingxian Li and Jiaji Liu},
      year={2026},
      eprint={2602.06718},
      archivePrefix={arXiv},
      primaryClass={cs.CR},
      url={https://arxiv.org/abs/2602.06718}, 
}

@misc{bragg2026astabench,
      title={AstaBench: Rigorous Benchmarking of AI Agents with a Scientific Research Suite}, 
      author={Jonathan Bragg and Mike D'Arcy and Nishant Balepur and Dan Bareket and Bhavana Dalvi and Sergey Feldman and Dany Haddad and Jena D. Hwang and Peter Jansen and Varsha Kishore and Bodhisattwa Prasad Majumder and Aakanksha Naik and Sigal Rahamimov and Kyle Richardson and Amanpreet Singh and Harshit Surana and Aryeh Tiktinsky and Rosni Vasu and Guy Wiener and Chloe Anastasiades and Stefan Candra and Jason Dunkelberger and Dan Emery and Rob Evans and Malachi Hamada and Regan Huff and Rodney Kinney and Matt Latzke and Jaron Lochner and Ruben Lozano-Aguilera and Cecile Nguyen and Smita Rao and Amber Tanaka and Brooke Vlahos and Peter Clark and Doug Downey and Yoav Goldberg and Ashish Sabharwal and Daniel S. Weld},
      year={2026},
      eprint={2510.21652},
      archivePrefix={arXiv},
      primaryClass={cs.AI},
      url={https://arxiv.org/abs/2510.21652}, 
}

@misc{chen2025mlrbench,
      title={MLR-Bench: Evaluating AI Agents on Open-Ended Machine Learning Research}, 
      author={Hui Chen and Miao Xiong and Yujie Lu and Wei Han and Ailin Deng and Yufei He and Jiaying Wu and Yibo Li and Yue Liu and Bryan Hooi},
      year={2025},
      eprint={2505.19955},
      archivePrefix={arXiv},
      primaryClass={cs.LG},
      url={https://arxiv.org/abs/2505.19955}, 
}

@misc{garikaparthi2026researchgym,
      title={ResearchGym: Evaluating Language Model Agents on Real-World AI Research}, 
      author={Aniketh Garikaparthi and Manasi Patwardhan and Arman Cohan},
      year={2026},
      eprint={2602.15112},
      archivePrefix={arXiv},
      primaryClass={cs.AI},
      url={https://arxiv.org/abs/2602.15112}, 
}

@misc{lupidi2026airsbench,
      title={AIRS-Bench: a Suite of Tasks for Frontier AI Research Science Agents}, 
      author={Alisia Lupidi and Bhavul Gauri and Thomas Simon Foster and Bassel Al Omari and Despoina Magka and Alberto Pepe and Alexis Audran-Reiss and Muna Aghamelu and Nicolas Baldwin and Lucia Cipolina-Kun and Jean-Christophe Gagnon-Audet and Chee Hau Leow and Sandra Lefdal and Hossam Mossalam and Abhinav Moudgil and Saba Nazir and Emanuel Tewolde and Isabel Urrego and Jordi Armengol Estape and Amar Budhiraja and Gaurav Chaurasia and Abhishek Charnalia and Derek Dunfield and Karen Hambardzumyan and Daniel Izcovich and Martin Josifoski and Ishita Mediratta and Kelvin Niu and Parth Pathak and Michael Shvartsman and Edan Toledo and Anton Protopopov and Roberta Raileanu and Alexander Miller and Tatiana Shavrina and Jakob Foerster and Yoram Bachrach},
      year={2026},
      eprint={2602.06855},
      archivePrefix={arXiv},
      primaryClass={cs.AI},
      url={https://arxiv.org/abs/2602.06855}, 
}

@misc{wang2026firebench,
      title={FIRE-Bench: Evaluating AI Agents on the Rediscovery of Scientific Insights}, 
      author={Zhen Wang and Fan Bai and Zhongyan Luo and Jinyan Su and Kaiser Sun and Xinle Yu and Jieyuan Liu and Kun Zhou and Claire Cardie and Mark Dredze and Zhiting Hu and Eric P. Xing},
      year={2026},
      eprint={2602.02905},
      archivePrefix={arXiv},
      primaryClass={cs.AI},
      url={https://arxiv.org/abs/2602.02905}, 
}

@misc{weng2025deepscientist,
      title={DeepScientist: Advancing Frontier-Pushing Scientific Findings Progressively}, 
      author={Yixuan Weng and Minjun Zhu and Qiujie Xie and Qiyao Sun and Zhen Lin and Sifan Liu and Yue Zhang},
      year={2025},
      eprint={2509.26603},
      archivePrefix={arXiv},
      primaryClass={cs.CL},
      url={https://arxiv.org/abs/2509.26603}, 
}

@misc{wu2025innovatorbench,
      title={InnovatorBench: Evaluating Agents' Ability to Conduct Innovative LLM Research}, 
      author={Yunze Wu and Dayuan Fu and Weiye Si and Zhen Huang and Mohan Jiang and Keyu Li and Shijie Xia and Jie Sun and Tianze Xu and Xiangkun Hu and Pengrui Lu and Xiaojie Cai and Lyumanshan Ye and Wenhong Zhu and Yang Xiao and Pengfei Liu},
      year={2025},
      eprint={2510.27598},
      archivePrefix={arXiv},
      primaryClass={cs.AI},
      url={https://arxiv.org/abs/2510.27598}, 
}

@misc{moussa2025scholareval,
      title={ScholarEval: Research Idea Evaluation Grounded in Literature}, 
      author={Hanane Nour Moussa and Patrick Queiroz Da Silva and Daniel Adu-Ampratwum and Alyson East and Zitong Lu and Nikki Puccetti and Mingyi Xue and Huan Sun and Bodhisattwa Prasad Majumder and Sachin Kumar},
      year={2026},
      eprint={2510.16234},
      archivePrefix={arXiv},
      primaryClass={cs.AI},
      url={https://arxiv.org/abs/2510.16234}, 
}

@misc{gao2025mmreview,
      title={MMReview: A Multidisciplinary and Multimodal Benchmark for LLM-Based Peer Review Automation}, 
      author={Xian Gao and Jiacheng Ruan and Zongyun Zhang and Jingsheng Gao and Ting Liu and Yuzhuo Fu},
      year={2025},
      eprint={2508.14146},
      archivePrefix={arXiv},
      primaryClass={cs.CL},
      url={https://arxiv.org/abs/2508.14146}, 
}

@inproceedings{kwa2025longtasks,
  author       = {Thomas Kwa and
                  Ben West and
                  Joel Becker and
                  Amy Deng and
                  Katharyn Garcia and
                  Max Hasin and
                  Sami Jawhar and
                  Megan Kinniment and
                  Nate Rush and
                  Sydney von Arx and
                  Ryan Bloom and
                  Thomas Broadley and
                  Haoxing Du and
                  Brian Goodrich and
                  Nikola Jurkovic and
                  Luke Harold Miles and
                  Seraphina Nix and
                  Tao Lin and
                  Neev Parikh and
                  David Rein and
                  Lucas Jun Koba Sato and
                  Hjalmar Wijk and
                  Daniel M. Ziegler and
                  Elizabeth Barnes and
                  Lawrence Chan},
  editor       = {Danielle Belgrave and
                  Cheng Zhang and
                  Laura N. Montoya and
                  Hsuan{-}Tien Lin and
                  Razvan Pascanu and
                  Piotr Koniusz and
                  Marzyeh Ghassemi and
                  Nancy Chen and
                  Iv{\'{a}}n Vladimir Meza Ru{\'{\i}}z and
                  Arturo Loaiza{-}Bonilla},
  title        = {Measuring {AI} Ability to Complete Long Software Tasks},
  booktitle    = {Advances in Neural Information Processing Systems 38: Annual Conference
                  on Neural Information Processing Systems 2025, NeurIPS 2025, San Diego,
                  CA, USA, December 2-7, 2025 / Mexico City, Mexico, November 30 - December
                  5, 2025},
  year         = {2025},
  url          = {http://papers.nips.cc/paper\_files/paper/2025/hash/85069585133c4c168c865e65d72e9775-Abstract-Conference.html},
  bibsource    = {dblp computer science bibliography, https://dblp.org}
}

@misc{lu2024aiscientist,
      title={The AI Scientist: Towards Fully Automated Open-Ended Scientific Discovery}, 
      author={Chris Lu and Cong Lu and Robert Tjarko Lange and Jakob Foerster and Jeff Clune and David Ha},
      year={2024},
      eprint={2408.06292},
      archivePrefix={arXiv},
      primaryClass={cs.AI},
      url={https://arxiv.org/abs/2408.06292}, 
}

@inproceedings{zheng2023mtbench,
  author       = {Lianmin Zheng and
                  Wei{-}Lin Chiang and
                  Ying Sheng and
                  Siyuan Zhuang and
                  Zhanghao Wu and
                  Yonghao Zhuang and
                  Zi Lin and
                  Zhuohan Li and
                  Dacheng Li and
                  Eric P. Xing and
                  Hao Zhang and
                  Joseph E. Gonzalez and
                  Ion Stoica},
  editor       = {Alice Oh and
                  Tristan Naumann and
                  Amir Globerson and
                  Kate Saenko and
                  Moritz Hardt and
                  Sergey Levine},
  title        = {Judging LLM-as-a-Judge with MT-Bench and Chatbot Arena},
  booktitle    = {Advances in Neural Information Processing Systems 36: Annual Conference
                  on Neural Information Processing Systems 2023, NeurIPS 2023, New Orleans,
                  LA, USA, December 10 - 16, 2023},
  year         = {2023},
  url          = {http://papers.nips.cc/paper\_files/paper/2023/hash/91f18a1287b398d378ef22505bf41832-Abstract-Datasets\_and\_Benchmarks.html},
  bibsource    = {dblp computer science bibliography, https://dblp.org}
}

@article{zhang2025exploring,
	title = {Exploring the role of large language models in the scientific method: from hypothesis to discovery},
	volume = {1},
	issn = {3005-1460},
	url = {https://doi.org/10.1038/s44387-025-00019-5},
	doi = {10.1038/s44387-025-00019-5},
	number = {1},
	journal = {npj Artificial Intelligence},
	author = {Zhang, Yanbo and Khan, Sumeer A. and Mahmud, Adnan and Yang, Huck and Lavin, Alexander and Levin, Michael and Frey, Jeremy and Dunnmon, Jared and Evans, James and Bundy, Alan and Dzeroski, Saso and Tegner, Jesper and Zenil, Hector},
	month = aug,
	year = {2025},
	pages = {14},
}

@misc{luo2025llm4sr,
      title={LLM4SR: A Survey on Large Language Models for Scientific Research}, 
      author={Ziming Luo and Zonglin Yang and Zexin Xu and Wei Yang and Xinya Du},
      year={2025},
      eprint={2501.04306},
      archivePrefix={arXiv},
      primaryClass={cs.CL},
      url={https://arxiv.org/abs/2501.04306}, 
}

@misc{chen2025ai4research,
      title={AI4Research: A Survey of Artificial Intelligence for Scientific Research}, 
      author={Qiguang Chen and Mingda Yang and Libo Qin and Jinhao Liu and Zheng Yan and Jiannan Guan and Dengyun Peng and Yiyan Ji and Hanjing Li and Mengkang Hu and Yimeng Zhang and Yihao Liang and Yuhang Zhou and Jiaqi Wang and Zhi Chen and Wanxiang Che},
      year={2025},
      eprint={2507.01903},
      archivePrefix={arXiv},
      primaryClass={cs.CL},
      url={https://arxiv.org/abs/2507.01903}, 
}

@inproceedings{zheng2025automation,
  author       = {Tianshi Zheng and
                  Zheye Deng and
                  Hong Ting Tsang and
                  Weiqi Wang and
                  Jiaxin Bai and
                  Zihao Wang and
                  Yangqiu Song},
  editor       = {Christos Christodoulopoulos and
                  Tanmoy Chakraborty and
                  Carolyn Rose and
                  Violet Peng},
  title        = {From Automation to Autonomy: {A} Survey on Large Language Models in
                  Scientific Discovery},
  booktitle    = {Proceedings of the 2025 Conference on Empirical Methods in Natural
                  Language Processing, {EMNLP} 2025, Suzhou, China, November 4-9, 2025},
  pages        = {17733--17750},
  publisher    = {Association for Computational Linguistics},
  year         = {2025},
  url          = {https://doi.org/10.18653/v1/2025.emnlp-main.895},
  doi          = {10.18653/V1/2025.EMNLP-MAIN.895},
  bibsource    = {dblp computer science bibliography, https://dblp.org}
}

@misc{wei2025ai,
      title={From AI for Science to Agentic Science: A Survey on Autonomous Scientific Discovery}, 
      author={Jiaqi Wei and Yuejin Yang and Xiang Zhang and Yuhan Chen and Xiang Zhuang and Zhangyang Gao and Dongzhan Zhou and Guangshuai Wang and Zhiqiang Gao and Juntai Cao and Zijie Qiu and Ming Hu and Chenglong Ma and Shixiang Tang and Junjun He and Chunfeng Song and Xuming He and Qiang Zhang and Chenyu You and Shuangjia Zheng and Ning Ding and Wanli Ouyang and Nanqing Dong and Yu Cheng and Siqi Sun and Lei Bai and Bowen Zhou},
      year={2025},
      eprint={2508.14111},
      archivePrefix={arXiv},
      primaryClass={cs.LG},
      url={https://arxiv.org/abs/2508.14111}, 
}

@misc{xie2025far,
      title={How Far Are AI Scientists from Changing the World?}, 
      author={Qiujie Xie and Yixuan Weng and Minjun Zhu and Fuchen Shen and Shulin Huang and Zhen Lin and Jiahui Zhou and Zilan Mao and Zijie Yang and Linyi Yang and Jian Wu and Yue Zhang},
      year={2025},
      eprint={2507.23276},
      archivePrefix={arXiv},
      primaryClass={cs.AI},
      url={https://arxiv.org/abs/2507.23276}, 
}

@misc{ren2025towards,
      title={Towards Scientific Intelligence: A Survey of LLM-based Scientific Agents}, 
      author={Shuo Ren and Can Xie and Pu Jian and Zhenjiang Ren and Chunlin Leng and Jiajun Zhang},
      year={2026},
      eprint={2503.24047},
      archivePrefix={arXiv},
      primaryClass={cs.AI},
      url={https://arxiv.org/abs/2503.24047}, 
}
\clearpage
\appendix
\captionsetup{hypcap=false}

\section{Comparative Mapping and Reporting Recommendations}
\label{sec:summary-tables}
Table~\ref{tab:compact-mapping} summarizes selected applications of the claim-strength analysis. Table~\ref{tab:reporting-template} records the task-dependent reporting recommendations discussed in Section~\ref{sec:agenda}.

\begin{strip}
\begin{minipage}{\textwidth}
\vspace*{12pt}
\small
\centering
\setlength{\tabcolsep}{3pt}
\renewcommand{\arraystretch}{1.12}
\begin{tabularx}{\textwidth}{C{0.16\textwidth}YYY}
\toprule
\textbf{Task / representative benchmark} & \textbf{Evidence, evaluator, and metric} & \textbf{Diagnostic information} & \textbf{Supported interpretation and conditions} \\
\midrule
Survey assessment / SurveyBench \citep{sun2025surveybench} & Human reference surveys and generated text; LLM content scores and quizzes answered from retrieved passages & Compares content scores with question answerability & \textbf{L2.} Reference-relative scores and simulated information use on 20 topics; human-written references do not themselves validate agreement with reader judgments \\
Idea assessment / AI Idea Bench \citep{qiu2025ai} & Target papers and retrieved literature; LLM matching judgments and citation-impact feasibility proxies & Separates reference matching from a literature-impact proxy for feasibility & \textbf{L2.} Reference-conditioned idea matching; citation impact characterizes associated literature, and the feasibility proxy is not calibrated to step applicability or compatibility \\
ML engineering / MLE-bench \citep{chan2025mlebench} & Competition data and submissions; local grader, historical medal thresholds, and pass@$k$ & Measures task completion and performance at stated attempt budgets & \textbf{L2.} ML-engineering task performance under the specified environment, resources, and attempt budget; historical medal thresholds provide a calibrated human-performance reference, while resplit test sets and reimplemented graders limit direct comparison with the original competitions \\
Related-work writing / GREP \citep{sahinuc2025expert} & Source papers, author preferences, generated sections, and revisions; criterion-specific checks and expert pairwise judgments & Separates source-support constraints from preferences about length, emphasis, and positioning; tests feedback-driven revision & \textbf{L4.} Expert-aligned assessment of related-work writing in the tested NLP/CV settings, alongside constraint checks; whole-section faithfulness scores include citation-parsing failures \\
Generated-review assessment / ReviewEval \citep{garg2025revieweval} & Manuscripts, reference reviews, venue rules, and generated reviews; alignment, factual-support, depth, constructiveness, and guideline scores & Compares dimensions of review quality; checks comments through manuscript retrieval and generated rebuttals & \textbf{L2.} Multidimensional review assessment under manuscript and reference checks; supplementary expert preferences on 30 papers do not independently calibrate the factual-support verifier \\
Integrated execution / AstaBench (E2E) \citep{bragg2026astabench} & Reports, code, and artifacts; rubric-based step scores and whole-task completion & Cross-checks reported work against execution artifacts & \textbf{L3.} Completion of specified research tasks with research-step verification; artifact checks expose discrepancies missed by report-only inspection \\
\bottomrule
\end{tabularx}
\captionof{table}{Selected applications of the claim-strength analysis in Figure~\ref{fig:claim-strength}. Labels L2--L4 summarize the core evaluation design described in each row. Scope conditions and identified support gaps are distinguished in the text.}
\label{tab:compact-mapping}
\end{minipage}
\end{strip}

\clearpage
\begin{strip}
\begin{minipage}{\textwidth}
\small
\centering
\setlength{\tabcolsep}{3pt}
\renewcommand{\arraystretch}{1.12}
\begin{tabularx}{\textwidth}{C{0.12\textwidth}C{0.15\textwidth}YYY}
\toprule
\textbf{Evaluation target} & \textbf{When applicable} & \textbf{Evidence and checks} & \textbf{Protocol details to report} & \textbf{Purpose} \\
\midrule
Literature synthesis & Source coverage and organization & Retrieved and supplied papers, reference structures, generated synthesis, and source-support checks & Retrieval cutoff and depth, reference construction, valid alternatives, and scoring denominators & Locate losses in retrieval, selection, or synthesis; interpret reference coverage \\
Literature synthesis & Answerability or reader benefit & Questions, answer evidence, reader tasks, and reference or control conditions & Human versus simulated readers, evidence access, participant sample, and aggregation & Distinguish supported answerability from observed reader benefit \\
Research ideation & Novelty or feasibility assessment & Prior-work comparisons, decomposed proposals, expert labels, and proxy-calibration results & Literature cutoff, retrieval coverage, judgment criteria, and reference-label construction & Identify what novelty and feasibility judgments are supported by \\
Research ideation & Proposal feedback or user benefit & Feedback requirements, evidence-bearing reports, and researcher judgments or revisions & Expert selection, blinding, agreement, comparison conditions, and measured outcomes & Relate feedback coverage to perceived usefulness or observed improvement \\
Executable workflows & Code or task correctness & Submitted programs, output tests, reference solutions, and execution results & Inputs, supplied decomposition, tests, environment, and permitted feedback & Separate successful execution from task-specific correctness \\
Executable workflows & Analysis decisions or performance comparison & Checked transformations and model choices; task scores and human attempts where available & Tool access, runtime, compute, retries, selection rules, and reference decisions & Explain analysis choices and make resource-conditioned comparisons interpretable \\
Scholarly writing & Section generation and revision & Source texts, author constraints, generated drafts, and criterion-specific judgments & Input sufficiency, parsing rules, rubric, evaluator calibration, and revision rounds & Separate content support, preference satisfaction, and formatting failures \\
Scholarly writing & Citation verification or visual communication & Source passages and validity labels for citations; source content, visual artifacts, and reader checks for communication & For verification: access limits and unavailable-source handling; for communication: fidelity criteria and reader conditions & Make citation decisions traceable; assess preserved content and accessibility \\
Automatic peer review & Generated-review quality & Manuscript, generated review, aspect scores, and independent judgments for the assessed dimensions & Evaluator, rubric, model/prompt, blinding, agreement, and calibration for each dimension & Distinguish critique support, analytical depth, constructiveness, and preference \\
Automatic peer review & Error discovery or feedback use & Reference errors and matched findings for detection; assignments, revisions, and outcomes for feedback & For detection: error construction and unmatched findings; for use: participants, delivery, uptake, and controls & Interpret detection coverage separately from observed workflow effects \\
End-to-end research & Connected research execution & Stage outputs, code, logs, reports, and checked experimental artifacts & Starting resources, handoffs, execution checks, tools, time, compute, and retries & Verify reported work and attribute failures across stages \\
End-to-end research & Overall completion or suite aggregation & Required steps, final outcomes, subtask scores, and missing or failed runs & Connected-run versus suite design, weights, denominators, and resource allocation & Separate whole-task completion from average component performance \\
\bottomrule
\end{tabularx}
\captionof{table}{Reporting recommendations organized by evaluation target and setting. Common fields are stated in Section~\ref{sec:agenda}. Requirements apply to the named setting; a cross-target protocol may use several rows.}
\label{tab:reporting-template}
\end{minipage}
\end{strip}

\clearpage

\section{Search and Protocol Records}
\label{sec:protocol-records}
The following tables document protocols within the review's six targets, including the studies discussed in the main text. Rows distinguish task inputs, the evidence used to judge outputs, scoring, calibration or controls, and findings or applicability conditions. Multiple protocols from one work are distinguished when their evidence or scoring differs.

Search terms combined each target with evaluation-related variants. Literature synthesis queries included survey generation, cross-paper comparison, and survey evaluation; ideation queries included research idea evaluation and novelty assessment; workflow queries included scientific code, data analysis, and ML research agents. Writing queries included citation-text generation, citation support, and scientific communication artifacts; reviewing queries included grounded critique, error localization, and review utility; end-to-end queries included research agents, research environments, and scientific rediscovery. Backward and forward citation tracing connected these searches. The tables preserve relevant alternative designs rather than all neighboring application areas.

\subsection{Scientific Literature Synthesis Protocols}
The protocols below distinguish survey generation, structured synthesis, and question-driven information use. Separate rows identify different scoring tasks within a work; human reference artifacts and independent human judgments are recorded separately.

\begin{strip}
\begin{minipage}{\textwidth}
\begingroup
\small
\setlength{\tabcolsep}{3pt}
\renewcommand{\arraystretch}{1.15}
\centering
\begin{tabular}{@{}*{5}{>{\raggedright\arraybackslash}p{0.18\textwidth}}@{}}
\toprule
\textbf{Work and task} & \textbf{Input/output evidence} & \textbf{Evaluator/metric} & \textbf{Calibration/controls} & \textbf{Findings/conditions}\\
\midrule
\multicolumn{5}{l}{\textbf{Survey generation and content assessment}}\\
AutoSurvey: survey quality \citep{wang2024autosurvey} &
Topic and retrieved papers; generated sections and citation passages. &
Multiple LLMs score coverage, structure, relevance, and citation support. &
Same writer/reference budget for naive retrieval baseline; three doctoral raters rank 20 surveys. &
Citation recall covers already-cited sentences; ranking correlation with humans is 0.5429.\\
AutoSurvey: answerability \citep{wang2024autosurvey} &
Survey passages support 50 multiple-choice questions from five topics. &
LLM question-answering accuracy. &
No material, naive retrieval survey, and option-specific retrieval conditions. &
AutoSurvey improves accuracy by 9.2 points over no material, 2.4 over naive retrieval; simulated use.\\
SurGE \citep{su2025surge} &
205 human surveys; shared title/abstract retrieval pool; generated references, outlines, and sections. &
Reference recall; document/section/sentence citation judgments; structure/content scores. &
Expert anonymous rankings calibrate structure/content; shared backbone for dedicated generators. &
Citation support uses abstracts; ranking calibration does not cover citation judgments.\\
SurveyGen/QUAL-SG \citep{bao2025surveygen} &
Topic-only, automatic retrieval, or supplied human references/outline; generated surveys. &
Reference authenticity and overlap; key-point recall, text/structure similarity, LLM ratings. &
120 surveys in four fields; three researchers compare 15 CS outputs with human originals. &
Resource contains 4,205 surveys, not 4,205 evaluated outputs; supplied outlines change the task.\\
Tang et al.: bibliography \citep{tangTangXueMei2025large} &
Title/keywords of Annual Reviews articles; ten generated references per query. &
Semantic Scholar metadata matching; authenticity precision and reference overlap. &
Three raters check 100 references; automated accuracy 86\%, $\kappa=0.71$. &
Matching allows partial metadata agreement; authenticity and overlap have different meanings.\\
\bottomrule
\end{tabular}
\captionof{table}{Detailed protocols for scientific literature synthesis.}\label{tab:appendix-literature-synthesis}
\endgroup
\end{minipage}
\end{strip}
\clearpage
\begin{strip}
\begin{minipage}{\textwidth}
\begingroup
\small
\setlength{\tabcolsep}{3pt}
\renewcommand{\arraystretch}{1.15}
\centering
\begin{tabular}{@{}*{5}{>{\raggedright\arraybackslash}p{0.18\textwidth}}@{}}
\toprule
\textbf{Work and task} & \textbf{Input/output evidence} & \textbf{Evaluator/metric} & \textbf{Calibration/controls} & \textbf{Findings/conditions}\\
\midrule
Tang et al.: abstracts \citep{tangTangXueMei2025large} &
Title/keywords; generated and human abstracts of matched length. &
Semantic similarity, overlap, and two textual-entailment judges. &
Five-model comparison; content reference is the human abstract. &
Published entailment formula tests generated-to-reference direction, not all extra generated assertions.\\
Tang et al.: short reviews \citep{tangTangXueMei2025large} &
Title, keywords, and human abstract; approximately 1,000-word review with ten citations. &
Bibliographic checks, overlap, and recall of ten reference key points. &
Same 1,105-review source collection; tasks are not a chained pipeline. &
Human abstract is supplied; later GPT-5 supplement uses a different 2025 collection.\\
SurveyForge \citep{yan2025surveyforge} &
Topic, retrieved papers, and human outline examples; survey and bibliography. &
Reference-set overlap; LLM outline/content scores. &
20 doctoral researchers blind-compare 100 outputs with AutoSurvey; same-backbone comparisons. &
Human outline preference is 74\% against AutoSurvey, not against human surveys; reference formula is precision-like.\\
SurveyEval \citep{zhao2025surveyeval} &
38 topics; human and generated surveys; reference lists. &
Human-anchored content rubric; reference-free outline rubric; bibliography precision/recall/F1. &
Common title queries; reference-list evaluation only for two of seven systems. &
Human text is an anchor, not independent human calibration; overlap does not identify hallucinated references.\\
SGSimEval \citep{guo2025sgsimeval} &
80 human surveys and collected system outputs; outline/content/reference embeddings. &
LLM ratings with similarity weighting toward full marks or measured human-text scores. &
Alternative LLM pairwise comparisons; no direct expert-score calibration reported. &
Some systems only cover CS topics; weighted scores describe unchanged outputs under another scale.\\
DeepSurvey-Bench: rubric \citep{zhang2026deepsurveybench} &
178 selected human surveys; generated surveys; seven academic-value criteria. &
Three-model ratings; human agreement and correlations with surface-quality scores. &
Three human raters assess 40 generated surveys; separate reference-selection annotation. &
Rubric and surface scores correlate only partly in this sample.\\
DeepSurvey-Bench: evidence audit \citep{zhang2026deepsurveybench} &
400 claim--evidence units from 40 high-scoring surveys, checked against sources. &
Human factual correctness, evidence alignment, and citation reliability. &
Separate from the rubric-calibration protocol; selection is conditional on high scores. &
Reported values are 94\%, 91\%, and 93\%; these are not whole-corpus accuracy estimates.\\
DeepSurvey-Bench: feedback \citep{zhang2026deepsurveybench} &
40 low-scoring surveys revised once with rubric feedback. &
Same model ensemble scores before/after on seven criteria. &
Topic, outline, scope, and reference set held fixed. &
Higher scores are reported; independent human assessment of revision gains is not reported.\\
\midrule
\multicolumn{5}{l}{\textbf{Comparison tables and field organization}}\\
ArxivDIGESTables: columns \citep{newman2024arxivdigestables} &
Titles/abstracts and optional captions/context; generated comparison dimensions. &
Semantic column alignment and schema recall; human usefulness, specificity, and insightfulness. &
50-table alignment audit; generated column count fixed to reference count. &
Alignment precision 70--85\%, depending on partial-match treatment; unmatched columns can be useful.\\
\bottomrule
\end{tabular}
\par\smallskip\noindent Table~\ref{tab:appendix-literature-synthesis} (continued): Detailed protocols for scientific literature synthesis.\par
\endgroup
\end{minipage}
\end{strip}
\clearpage
\begin{strip}
\begin{minipage}{\textwidth}
\begingroup
\small
\setlength{\tabcolsep}{3pt}
\renewcommand{\arraystretch}{1.15}
\centering
\begin{tabular}{@{}*{5}{>{\raggedright\arraybackslash}p{0.18\textwidth}}@{}}
\toprule
\textbf{Work and task} & \textbf{Input/output evidence} & \textbf{Evaluator/metric} & \textbf{Calibration/controls} & \textbf{Findings/conditions}\\
\midrule
ArxivDIGESTables: cells \citep{newman2024arxivdigestables} &
Human schema and paper full text; generated cell values. &
Automatic semantic matching; human full/partial/no-match labels. &
30-table audit, two raters per value; context variants. &
Extra context mainly improves partial matches; this is conditional filling, not end-to-end table accuracy.\\
arXiv2Table \citep{wang2025arxiv2table} &
Rewritten request, papers and annotated distractors; selected papers and comparison table. &
Paper-selection precision/recall/F1; bidirectional questions on schema, cells, and cell pairs. &
Human checks of 200 questions and 300 judgments per method; judge-swap controls. &
Main tasks use candidate pools; novel content beyond the reference receives no credit by design.\\
TaxoBench \citep{zhang2026can} &
Topic-only retrieval/organization or supplied expert paper set; hierarchy and paper assignments. &
Paper overlap, leaf clustering, semantic tree/path distances, and tree statistics. &
Random/depth controls; two independent human organizers on ten topics. &
Path scores depend on labels/depth; ten-topic human comparison is not a 72-topic estimate.\\
TaxoAlign/CS-TAXOBENCH \citep{lahiri2025taxoalign} &
Topic and supplied reference papers; generated classification tree; extracted human trees. &
Node recall, text similarity, size ratio, and model/human structure/content ratings. &
Three raters on 20 trees; reference-extraction audit; additional conference test set. &
References are supplied. Degree-sum ratio measures tree size, not parent--child alignment.\\
\midrule
\multicolumn{5}{l}{\textbf{Question-driven synthesis and information use}}\\
OpenScholar: ScholarQABench \citep{asai2024openscholar} &
Scientific questions and open retrieval; cited answers, expert answers, and rubrics. &
Automatic answer/citation scores; expert quality, usefulness, and pairwise preference. &
108 questions, 16 experts; no self-question rating; 50-question length control. &
Expert preference and citation correctness are separate; experimental corpus is peS2o v2.\\
ResearchQA \citep{yifei2025researchqa} &
Survey-mined questions and requirements; long scholarly answers with citations. &
LLM requirement coverage, pairwise preference, and tournament ranking. &
31 doctoral-level annotators; validation in eight of 75 fields; length sensitivity on 225 questions. &
Coverage is recall-like; incomplete citation requirements do not imply fabricated citations.\\
SurveyBench: content \citep{sun2025surveybench} &
20 topics, human references, generated surveys and outlines. &
LLM outline/content rubrics; non-text-element density. &
Reference versus no-reference scoring; human quality anchors fixed to full marks. &
DeepResearch content mean 4.42/5; anchor-based scores are not independent expert ratings.\\
SurveyBench: quizzes \citep{sun2025surveybench} &
General/topic questions; passages retrieved from evaluated survey; cited answers. &
LLM answer quality and evidence checks; unsupported answer scored zero. &
Human-survey answering reference; topic-specific anchor fixed to 10. &
Topic-specific scores 1.48--3.19/10 across four systems; measures simulated supported answerability.\\
Deep Research, Shallow Evaluation \citep{hwang2026deep} &
Scientific reports; holistic preferences and criterion-specific expert annotations. &
Agreement and rank correlation at system and instance levels. &
200-question holistic study; separate near/deep-expert studies; 80 fixed-claim citation checks. &
System agreement can conceal weaker instance agreement; expertise comparisons also change questions and system pool.\\
MDA-QA \citep{huang2025multidocument} &
Citation-linked paper communities; 6,804 filtered questions requiring multiple sources. &
Support-document recall/exact-match retrieval; answer similarity metrics. &
Single-document answer screening; separate 30- and 200-question expert checks. &
198/200 audited questions meet all criteria; automatic filtering is not a strict necessity proof for every item.\\
\bottomrule
\end{tabular}
\par\smallskip\noindent Table~\ref{tab:appendix-literature-synthesis} (continued): Detailed protocols for scientific literature synthesis.\par
\endgroup
\end{minipage}
\end{strip}
\clearpage
\begin{strip}
\begin{minipage}{\textwidth}
\begingroup
\small
\setlength{\tabcolsep}{3pt}
\renewcommand{\arraystretch}{1.15}
\centering
\begin{tabular}{@{}*{5}{>{\raggedright\arraybackslash}p{0.18\textwidth}}@{}}
\toprule
\textbf{Work and task} & \textbf{Input/output evidence} & \textbf{Evaluator/metric} & \textbf{Calibration/controls} & \textbf{Findings/conditions}\\
\midrule
PaperQA2/LitQA2 \citep{skarlinski2024language} &
248 full-text-evidence questions; retrieval traces and answers. &
Answer accuracy/precision and supporting-DOI recall. &
Nine doctoral researchers; three system runs; contaminated human answers excluded. &
Answer precision 85.2\%; accuracy 66.0\%, not significantly above human accuracy.\\
PaperQA2/WikiCrow \citep{skarlinski2024language} &
Generated gene entries paired with existing Wikipedia entries; cited statements. &
Human citation-support categories and precision. &
375 statement evaluations by four experts, blinded to source. &
Experts judge existing entries; they do not write matched new comparison articles.\\
PaperQA2: ContraCrow \citep{skarlinski2024language} &
Claims from 93 biology papers; retrieved contradictory evidence. &
Graded contradiction judgments and human candidate validation. &
100 high-scoring candidates audited; separate detection setting without system reasoning. &
70\% of sampled candidates validated; human validation uses supplied evidence, not unrestricted search.\\
\bottomrule
\end{tabular}
\par\smallskip\noindent Table~\ref{tab:appendix-literature-synthesis} (continued): Detailed protocols for scientific literature synthesis.\par
\endgroup
\end{minipage}
\end{strip}

\clearpage
\subsection{Research Ideation and Proposal Protocols}
Expert judgments, published-paper references, literature-relative novelty, and proposal feedback use different labels. The table separates assessment of ideas from assessment of interactive support or report usefulness.

\begin{strip}
\begin{minipage}{\textwidth}
\begingroup
\small
\setlength{\tabcolsep}{3pt}
\renewcommand{\arraystretch}{1.15}
\centering
\begin{tabular}{@{}*{5}{>{\raggedright\arraybackslash}p{0.18\textwidth}}@{}}
\toprule
\textbf{Work and task} & \textbf{Input/output evidence} & \textbf{Evaluator/metric} & \textbf{Calibration/controls} & \textbf{Findings/conditions}\\
\midrule
\multicolumn{5}{l}{\textbf{Expert judgments and reference-driven comparisons}}\\
Si et al. \citep{si2024can} &
Human, AI-selected, and human-reranked AI proposals; shared templates and seven NLP topics. &
Blinded expert novelty, excitement, feasibility, expected effectiveness, and overall ratings. &
104 distinct researchers; 49 proposals per condition; style normalization and multiple statistical units. &
Higher AI novelty ratings; feasibility differences not significant at idea level. AI uses extensive sampling.\\
AI Idea Bench: target matching \citep{qiu2025ai} &
3,495 target papers; inspiration papers/topic; generated motivation and plan. &
LLM multiple-choice identification, idea/topic matching, and pairwise ranking. &
Common generation backbone; two candidates with best-of-two aggregation; researchers screen inspiration papers. &
Motivation and plan rankings differ; one baseline lacks plans. Target matching is reference recovery.\\
AI Idea Bench: literature proxies \citep{qiu2025ai} &
Idea components linked to prior/contemporary papers and citation statistics. &
Literature-distance novelty score; citation-impact feasibility and per-step scores. &
Time weighting and citation normalization; no independent step-feasibility calibration reported. &
Citation influence does not check method applicability or step compatibility; in the original study's Eq.~(7), larger historical distance lowers the novelty score when other terms are fixed.\\
IdeaBench \citep{guo2024ideabench} &
Biomedical target-paper references; three generated ideas mixed with target idea. &
GPT-4o relative novelty/feasibility ranking; semantic similarity and overlap. &
Five filtered references versus all references; blinded origins; fixed candidate count. &
Insight Score is the fraction ranked above the target; authors report high scores for some incoherent outputs.\\
AP-FRI \citep{kumar2025can} &
1,250 papers with future-work text removed; generated directions and author-written references. &
Idea matching and semantic diversity; expert novelty, relevance, feasibility/correctness. &
Matcher checked on labeled pairs; author studies in CS and physics; retrieval comparison. &
Author alignment does not exhaust valid directions; the original study's Eq.~(1) sums over generated ideas, while the following text defines the denominator as the author's idea count. The five Claude novelty percentages in the original study's Section~5.3 sum to 110.41\%.\\
\bottomrule
\end{tabular}
\captionof{table}{Detailed protocols for research ideation and proposal evaluation.}\label{tab:appendix-ideation}
\endgroup
\end{minipage}
\end{strip}
\clearpage
\begin{strip}
\begin{minipage}{\textwidth}
\begingroup
\small
\setlength{\tabcolsep}{3pt}
\renewcommand{\arraystretch}{1.15}
\centering
\begin{tabular}{@{}*{5}{>{\raggedright\arraybackslash}p{0.18\textwidth}}@{}}
\toprule
\textbf{Work and task} & \textbf{Input/output evidence} & \textbf{Evaluator/metric} & \textbf{Calibration/controls} & \textbf{Findings/conditions}\\
\midrule
LiveIdeaBench \citep{ruan2026evaluating} &
Single scientific keywords; short ideas across 1,180 keywords in 22 disciplines. &
Model-panel originality, feasibility, clarity, diversity; lower-quantile aggregate. &
Self-judging prohibited; six differential-equations experts assess 22 ideas; dynamic judge selection. &
Domain calibration uses all eligible judges, unlike main random panels; quantile is keyword-weighted.\\
GraphEval: historical labels \citep{feng2025grapheval} &
Paper titles/abstracts, viewpoint graph, and known training labels. &
Label propagation or graph network predicts historical decision classes; accuracy/macro-F1. &
ICLR train/test split; alternative prompting/classifier baselines. &
Supervised reference is a decision label; AI Researcher baselines use differing test scopes.\\
GraphEval: derived ideas \citep{feng2025grapheval} &
Copied, replaced, or recombined viewpoints with time features. &
Classification of synthetic derivative ideas. &
80 constructed examples; ten used to create training negatives. &
Reported detection improvement concerns constructed derivatives; the original study's Section~7.2 does not state whether the ten ideas used to construct training negatives are excluded from testing.\\
\midrule
\multicolumn{5}{l}{\textbf{Literature-relative novelty and interactive support}}\\
Literature-Grounded Novelty Assessment \citep{shahid2025literaturegrounded} &
Idea and retrieved top-ten papers; expert examples and binary labels with reasons. &
Accuracy, precision/recall/F1, and Cohen's kappa. &
35 training/32 test ideas; common papers for prompt comparisons; separate 58 non-novel cases. &
Test accuracy 0.81, $\kappa=0.59$; 58-case result is restricted to non-novel ideas.\\
Navigating Ideation Space \citep{shen2026navigating} &
Problem, method, and findings representations; component-linked prior papers and expert idea scores. &
Retrieval ranking metrics; similarity-derived novelty correlation. &
466 valid retrieval queries; 93 expert-scored ideas; retrieval-budget comparisons. &
Spearman 0.384 versus GPT-5 Mini 0.385; does not establish superiority on all measures.\\
Scideator: user support \citep{radensky2026humanllm} &
Seed papers and selected problem/method/evaluation facets; interaction logs and questionnaires. &
Creativity Support Index, confidence, interviews, and paired tests. &
22 researchers, two tools for 20 minutes each; same LLM, different tool configurations. &
Median support 70.50 versus 61.00; not isolated facet-only effect.\\
Scideator: novelty module \citep{radensky2026humanllm} &
Retrieved papers, generated ideas, expert novelty examples and labels. &
Binary novelty accuracy/F1; known-non-novel retrieval ablations. &
Shared evidence with Literature-Grounded Novelty Assessment; reported split sizes conflict. &
The module table does not establish an independent replication. The original study's Section~2 reports 67 labeled items and 20 in-context examples per class, while the original study's Section~4.2.2 reports 43 held-out items; these counts do not specify a consistent disjoint split.\\
\bottomrule
\end{tabular}
\par\smallskip\noindent Table~\ref{tab:appendix-ideation} (continued): Detailed protocols for research ideation and proposal evaluation.\par
\endgroup
\end{minipage}
\end{strip}
\clearpage
\begin{strip}
\begin{minipage}{\textwidth}
\begingroup
\small
\setlength{\tabcolsep}{3pt}
\renewcommand{\arraystretch}{1.15}
\centering
\begin{tabular}{@{}*{5}{>{\raggedright\arraybackslash}p{0.18\textwidth}}@{}}
\toprule
\textbf{Work and task} & \textbf{Input/output evidence} & \textbf{Evaluator/metric} & \textbf{Calibration/controls} & \textbf{Findings/conditions}\\
\midrule
SciMON \citep{wang2024scimon} &
Background, optional seed, and literature inspirations; generated idea sentence and revisions. &
Expert helpfulness, technical detail, relative novelty; auxiliary text similarity. &
Separate generation and iteration studies; 194 human-cleaned test cases; preliminary biomedical study. &
Changed wording and increased novelty are separately judged; supplied seed/reference conditions are explicit.\\
\midrule
\multicolumn{5}{l}{\textbf{Proposal reports and their evaluation}}\\
ScholarEval: report assessment \citep{moussa2025scholareval} &
Problem, method, and experiment plans; full-text soundness evidence and abstract-based contribution comparisons. &
Coverage of 1,076 expert-checked requirements; report preferences; invalid-link rate. &
117 ideas, six rubric validators; dated retrieval; module ablations. &
Zero links failing specified HTTP checks does not establish absence of all citation-support errors.\\
ScholarEval: user study \citep{moussa2025scholareval} &
Researchers' ideas and blinded ScholarEval/deep-research reports. &
Useful citations and report-quality/refinement questionnaire scores; mixed-effects analysis. &
18 experts, 46 evaluations; different ideas used with each system. &
More useful references and higher refinement-related ratings in this study.\\
InnoEval \citep{qiao2026innoeval} &
Timestamped ideas, paper/web/code evidence, and simulated reviewer perspectives. &
Historical-label accuracy/F1, pair/group ranking, report preference, expert-score correlation. &
Common DeepSeek-V3.2 backbone; 60 expert-rated ideas; knowledge-time separation. &
Best-choice accuracy 65.12\%, exact ranking 22.09\%; distinct ranking endpoints.\\
\bottomrule
\end{tabular}
\par\smallskip\noindent Table~\ref{tab:appendix-ideation} (continued): Detailed protocols for research ideation and proposal evaluation.\par
\endgroup
\end{minipage}
\end{strip}

\clearpage
\subsection{Executable Workflow and Scientific Coding Protocols}
These protocols inspect scientific functions, analysis decisions, program outputs, or machine-learning improvements. Execution tests, model judgments, and process measurements remain separate; resource and feedback conditions specify how each result was obtained.

\begin{strip}
\begin{minipage}{\textwidth}
\begingroup
\small
\setlength{\tabcolsep}{3pt}
\renewcommand{\arraystretch}{1.15}
\centering
\begin{tabular}{@{}*{5}{>{\raggedright\arraybackslash}p{0.18\textwidth}}@{}}
\toprule
\textbf{Work and task} & \textbf{Input/output evidence} & \textbf{Evaluator/metric} & \textbf{Calibration/controls} & \textbf{Findings/conditions}\\
\midrule
\multicolumn{5}{l}{\textbf{Scientific code and analysis decisions}}\\
SciCode \citep{tian2024scicode} &
Scientist-defined functions and workflow; generated code, reference implementations, numerical/domain tests. &
Subproblem and main-problem pass@1; main success requires steps and integration. &
At least two domain reviewers per problem; optional background and gold-prior-step controls. &
65 main/288 subproblems in test set. Decomposition is provided, not independently generated.\\
BLADE: decision selection \citep{gu2024blade} &
Research question/data; 188 questions on most/least defensible analysis choices. &
Multiple-choice accuracy. &
Expert-derived reference decisions; deterministic generation setting. &
Measures decision recognition; separate from implementing an analysis.\\
BLADE: analysis generation \citep{gu2024blade} &
12 research questions/data sets; variables, transformation functions, and statistical-model functions. &
Semantic, executed-value, and dataflow matching; precision, coverage@10, combined F1. &
11 experts construct 536 decisions; two authors audit 615 evaluator outputs. &
GPT-4o produces runnable nonempty analyses in 96\% of cases, yet model/variable coverage stays below 13\%.\\
DiscoveryBench \citep{majumder2024discoverybench} &
Real/synthetic data, metadata, and constrained question; generated hypothesis. &
LLM-assisted matching of conditions, variables, and relationships, on a 0--100 scale. &
No-data baseline; separate oracle-score-feedback condition; real test set has 239 tasks. &
Llama-3 no-data/CodeGen scores 11.5/12.1; matching alone does not isolate data verification.\\
ScienceAgentBench \citep{chen2025scienceagentbench} &
Task instructions, data, optional expert knowledge; self-contained Python and saved outputs. &
Valid execution and task success; numerical/prediction checks, some visual model scoring; code rubric. &
102 tasks, nine domain experts; three attempts; hidden labels/revised data where applicable. &
Best-of-three is not single-run success; human code rubric can mix reference implementation with partial completion.\\
DSEval \citep{zhang2024benchmarking} &
Queries, reference code, files, and session state; code and updated runtime state. &
Configurable validators check results, presentation, side effects, and execution; pass rate. &
825 questions in 294 groups; default reference-state resets versus error propagation. &
Ignoring presentation/integrity changes the success definition; error propagation is separately measured.\\
DSAEval \citep{sun2026dsaeval} &
Persistent multimodal notebook sessions; data, soft references, reasoning, code, and reports. &
Two model judges weight reasoning/code/results by 0.3/0.3/0.4; cost and judge error. &
Three doctoral raters calibrate on Gemini outputs for all 641 questions; visual-observation ablation. &
Judge calibration covers one generator; code score is not hidden-test success.\\
\bottomrule
\end{tabular}
\captionof{table}{Detailed protocols for executable research workflows and scientific coding.}\label{tab:appendix-executable}
\endgroup
\end{minipage}
\end{strip}
\clearpage
\begin{strip}
\begin{minipage}{\textwidth}
\begingroup
\small
\setlength{\tabcolsep}{3pt}
\renewcommand{\arraystretch}{1.15}
\centering
\begin{tabular}{@{}*{5}{>{\raggedright\arraybackslash}p{0.18\textwidth}}@{}}
\toprule
\textbf{Work and task} & \textbf{Input/output evidence} & \textbf{Evaluator/metric} & \textbf{Calibration/controls} & \textbf{Findings/conditions}\\
\midrule
Curie \citep{kon2025curie} &
Computational experiment questions, constraints, code, plans, execution records, and conclusions. &
LLM design/setup/conclusion checks; human implementation--plan alignment. &
46 reported tasks; shared GPT-4o; five runs; expert judge check described without sample/agreement detail. &
Design pass 97.9\%, conclusion correctness 36.1\%; tasks generally do not require iterative hypothesis revision.\\
\midrule
\multicolumn{5}{l}{\textbf{Machine-learning experimentation and research engineering}}\\
MLAgentBench \citep{huang2023mlagentbench} &
13 tasks, workspaces, baselines, file/code tools; final predictions or timed code. &
Task-native metrics; success usually requires 10\% relative improvement; time/token measures. &
Eight runs; usually 50 actions/five hours, but GPT-4 has 30 actions. &
Improvement means use of specified baseline; appendix CIFAR example exposes test feedback during tuning.\\
MLE-bench \citep{chan2025mlebench} &
75 offline competitions; descriptions/data; prediction submissions, hidden-test scores, and logs. &
Local task graders, valid submissions, historical medal thresholds, and pass@k. &
24 hours/one A10 in main setup; sample-solution split checks; separate contamination/log audits. &
o1-preview/AIDE 16.9\% medal rate; pass@8 34.1\%. Reconstructed tests limit exact historical equivalence.\\
RE-Bench \citep{wijk2024rebench} &
Seven optimization environments, weak starts, protected graders, expert/agent code and timed traces. &
Task metrics normalized to start/reference solutions; best-score summaries and uncertainty. &
61 experts, 71 eight-hour attempts; matched environments and budget-based comparisons. &
32-hour human result is four independent attempts; selected noisy scores need re-evaluation for stable quality.\\
MLRC-Bench \citep{zhang2025mlrcbench} &
Seven research competitions; protected repositories; development-selected snapshots and hidden-test results. &
Improvement relative to baseline/human solution; time and logical code lines; separate novelty study. &
Usually eight attempts, 50 steps/five hours; task-specific test budgets; optional external ideas separated. &
Scores may be negative. Novelty/effectiveness correlation $-0.06$ does not validate novelty judgments against novelty truth.\\
MLGym \citep{nathani2025mlgym} &
13 tasks; code, weights, or policies; intermediate attempts and final submissions. &
Task metrics; relative-performance profiles and area under those curves for best-of-four outputs. &
Shared agent interface; four seeds; protected graders; repeated score feedback. &
Best attempt and best submission are separate aggregates, not per-run best-solution retention rates.\\
FML-bench \citep{zou2025fmlbench} &
Eight research repositories; protected graders; iteration code and performance traces. &
Task metrics; code-embedding dispersion; successful/executed and executed/allocated step rates. &
Three rounds of 100 allocated steps; multiple agent/model combinations; failures recorded. &
Code dispersion is not independently calibrated to hypothesis diversity; the original study's Table~2 ranks lower privacy scores as better, whereas its Appendix~A defines the target as attack AUC near 0.5. These criteria differ if the tabulated scores are untransformed AUC.\\
\bottomrule
\end{tabular}
\par\smallskip\noindent Table~\ref{tab:appendix-executable} (continued): Detailed protocols for executable research workflows and scientific coding.\par
\endgroup
\end{minipage}
\end{strip}

\clearpage
\subsection{Academic Writing and Scholarly Communication Protocols}
The records separate source-supported prose and aspect-level scoring, bibliographic validity and contextual citation support, and visual communication. Separate rows distinguish protocols within one work; calibration samples are not counted as additional works.

\begin{strip}
\begin{minipage}{\textwidth}
\begingroup
\small
\setlength{\tabcolsep}{2.2pt}
\renewcommand{\arraystretch}{1.15}
\centering
\begin{tabular}{@{}>{\raggedright\arraybackslash}p{0.15\textwidth}>{\raggedright\arraybackslash}p{0.21\textwidth}>{\raggedright\arraybackslash}p{0.18\textwidth}>{\raggedright\arraybackslash}p{0.17\textwidth}>{\raggedright\arraybackslash}p{0.19\textwidth}@{}}
\toprule
\textbf{Work and task} & \textbf{Input/output evidence} & \textbf{Evaluator/metric} & \textbf{Calibration/controls} & \textbf{Findings/conditions} \\
\midrule
\multicolumn{5}{@{}l@{}}{\textbf{Source-supported writing and aspect-level evaluation}} \\
CiteBench \citep{funkquist2022citebench}; citation text & Given cited papers and citing context; original citation sentences or paragraphs are references. & ROUGE, BERTScore, citation-intent/discourse KL; human readability and consistency. & Three annotators rate 150 outputs; 50 overlapping cases assess agreement. & Some reference targets require absent input facts. Authors document examples; affected prevalence is unquantified. \\
SciIG \citep{garg2025lets}; introductions & Target title/abstract and cited metadata; generated and original introductions. & Reference-set precision/recall, keyword overlap, text metrics, model and human rubrics. & 20 prompts, 80 outputs; three authors divide annotation without overlap. & 3,900 tasks. Keyword factuality proxies align weakly with human ratings; source-set recovery is separate. \\
Task exploration \citep{sahinuc2024systematic}; citation paragraphs & Abstracts, reference-derived intent, and citation examples; generated paragraphs and atomic reference facts. & Text similarity, NLI, surface compliance; human atomic-fact coverage. & Two models; six prompts and six input settings; three annotators, 30 human-study examples. & Input manipulation is explicit. Automatic metrics only partly track atomic-fact coverage in this exploratory study. \\
GREP \citep{sahinuc2025expert}; related-work constraints & Main/cited abstracts and introductions, author preferences; generated sections and revisions. & Hard citation, support and positioning checks; soft length, emphasis and positioning-style scores. & 50 balanced examples per criterion; 16 experts, three rounds each; separate synthetic and interactive checks. & Whole-section faithfulness requires every citation to pass. Parsing failures and semantic errors are not isolated in that rate. \\
OARelatedWork \citep{docekal2024oarelatedwork}; related-work generation & 94,450 tasks with linked sources; validation/test require full texts; generated sections and statements. & Text similarity, citation F1; True/Wrong Citation/Unverifiable/False proportions; preference scores. & 40 papers, 120 sections; 408 statements including adversarial cases; judge meta-evaluation. & Full-text input changes attribution errors. Unverifiable is not False; selected judge's wrong-citation F1 is .23 in the pre-extracted test. \\
STRoGeNS \citep{nishimura2024structured}; paragraph organization & Target/cited titles and abstracts; paragraphs treated as citation clusters. & Cluster F1, ARI/ARI', paragraph-count error, missing-citation rate, ROUGE. & Four people rate 25 Llama2 outputs; Fleiss' $\kappa=.156$. & Human calibration concerns whole-section structure; evidence for paragraph-level measurement remains limited. \\
STRoGeNS \citep{nishimura2024structured}; novelty statements & Generated related work; original and statement-deleted texts for validation. & GPT-4 detects statement presence; generation study samples 100 outputs. & 190 original/constructed cases; reported detection accuracy 92\%. & Measures presence of a novelty statement; authors leave its validity for future work. \\
SciXGen \citep{chen2021scixgen}; object descriptions & Scientific object plus preceding text; author prose describing tables, figures, algorithms or theorems. & BLEU, METEOR, MoverScore, perplexity; 1--3 human ratings. & Five CS annotators; 50 samples per task; object-only, local and retrieved-context controls. & Description targets are prose spans, not captions. Context benefits vary by object type. \\
\bottomrule
\end{tabular}
\captionof{table}{Academic writing and scholarly communication protocols.}\label{tab:appendix-writing}
\endgroup
\end{minipage}
\end{strip}
\clearpage
\begin{strip}
\begin{minipage}{\textwidth}
\begingroup
\small
\setlength{\tabcolsep}{2.2pt}
\renewcommand{\arraystretch}{1.15}
\centering
\begin{tabular}{@{}>{\raggedright\arraybackslash}p{0.15\textwidth}>{\raggedright\arraybackslash}p{0.21\textwidth}>{\raggedright\arraybackslash}p{0.18\textwidth}>{\raggedright\arraybackslash}p{0.17\textwidth}>{\raggedright\arraybackslash}p{0.19\textwidth}@{}}
\toprule
\textbf{Work and task} & \textbf{Input/output evidence} & \textbf{Evaluator/metric} & \textbf{Calibration/controls} & \textbf{Findings/conditions} \\
\midrule
SciXGen \citep{chen2021scixgen}; introduction paragraphs & Paper abstract and cited abstracts/full-text passages; original introduction prose. & Text metrics, human ratings; proportion of supplied references not mentioned. & Abstract-only versus cited-abstract/full-text inputs; CV subset. & The named `Hallucination' percentage measures unmentioned references, supporting a coverage conclusion rather than a factual-error rate. \\
SciRM \citep{sahinuc2026reward}; aspect scoring & Query, explicit rubric and score examples; related-work/review-utility aspect labels. & Accuracy/F1 and repeated-run variability for rule-conditioned scores. & Human and synthetic labels; masked aspects; first-stage versus reflection training. & 65,357 aspect-level instances, not independent papers. Four-task means weight tasks equally. \\
SciRM \citep{sahinuc2026reward}; unseen-task transfer & Novelty-assessment summary pairs and scientific revision examples. & Label accuracy/F1 on agreement and revision evaluation. & 76 novelty pairs, 56 agreement labels; 6,184 revision aspect instances. & Novelty agreement accuracy near .74 is close to the majority baseline; the original study's Table~13 reports F1 without specifying the positive class or class-averaging convention. \\
\multicolumn{5}{@{}l@{}}{\textbf{Bibliographic validity and citation support}} \\
Walters--Wilder \citep{walters2023fabrication}; generated references & 84 generated reviews on 42 topics; 636 references checked against databases/publishers. & Manual existence, substantive metadata, formatting and link checks. & April 2023 GPT-3.5/GPT-4 outputs; suspected fabrications receive further source checks. & Fabrication: 55\%/18\%; errors among real works: 43\%/24\%. These have different denominators. \\
Chelli et al. \citep{chelli2024hallucination}; review-search references & 11 shoulder-disease systematic reviews, 109 included references; 362 model-generated entries. & Manual metadata checks, hallucination proportion, retrieval precision/recall. & Three 2023 models; two prompts per task, retaining the larger reference output. & Hallucination requires two of title/first author/year to be wrong. GPT-4 recall count is inconsistent (15 versus 16). \\
CiteAudit \citep{shi2026citeauditb}; metadata verification & Generated set: 6,086 entries; real-world set: 3,356; search-backed valid/invalid labels. & Error-positive confusion matrices, accuracy, precision, recall and F1. & At least two authors independently verify labels; unresolved disagreements excluded. & Reported real-set precision is .749 and recall is .951. Similar recall for one tool does not establish equivalent error distributions across sets. \\
CiteTracer \citep{li2026source}; synthetic field verification & Perturbed BibTeX/PDF references; Real, Potential and Hallucinated codes and evidence links. & One-versus-rest F1; per-code true-/false-positive rates. & Triple validation; 2,450 BibTeX entries, 2,392 after PDF-rendering losses; five styles. & 11 of 12 taxonomy classes tested. P3's matched-real-work definition conflicts with its pure-fabrication construction description. \\
CiteTracer \citep{li2026source}; high-risk real references & 957 preselected positive references from rejected conference papers; adjudication evidence. & Recall of Hallucinated labels; Potential reported separately. & Conference-chair decisions support the selected real-world cases. & 929/957 labeled Hallucinated; flagging remaining cases as Potential is not the same classification endpoint. \\
Naser \citep{naser2026how}; reference audit & Non-retrieval outputs from ten models/four fields; 69,557 reported references. & Database matching, thresholded confirmation rates, repeated-output agreement. & CrossRef, OpenAlex and Semantic Scholar; automated search rechecks 75 cases per stratum. & Unconfirmed includes some real references. Authors treat the rate as an upper bound under this matching definition. \\
\bottomrule
\end{tabular}
\par\smallskip\noindent Table~\ref{tab:appendix-writing} (continued): Academic writing and scholarly communication protocols.\par
\endgroup
\end{minipage}
\end{strip}
\clearpage
\begin{strip}
\begin{minipage}{\textwidth}
\begingroup
\small
\setlength{\tabcolsep}{2.2pt}
\renewcommand{\arraystretch}{1.15}
\centering
\begin{tabular}{@{}>{\raggedright\arraybackslash}p{0.15\textwidth}>{\raggedright\arraybackslash}p{0.21\textwidth}>{\raggedright\arraybackslash}p{0.18\textwidth}>{\raggedright\arraybackslash}p{0.17\textwidth}>{\raggedright\arraybackslash}p{0.19\textwidth}@{}}
\toprule
\textbf{Work and task} & \textbf{Input/output evidence} & \textbf{Evaluator/metric} & \textbf{Calibration/controls} & \textbf{Findings/conditions} \\
\midrule
Naser \citep{naser2026how}; string-only detection & 27 bibliographic string features; confirmed/unconfirmed reference labels. & AUC, average precision and accuracy. & Five-fold stratified validation; leave-one-model-out evaluation. & GBM AUC .876; mean leave-one-model-out AUC .834. Labels inherit the audit's matching definition. \\
BibTeX errors \citep{rao2026bibtex}; reconstruction and repair & Known title/first author; 931 papers, 2,793 entries; nine fields and multi-version metadata. & Six field states; field/entry correctness and repair regressions. & 521-field human reannotation; native-search, one-stage and two-stage tool conditions. & The evaluation uses first-author matching; stricter author checks differ. Authors report loss of original data, preventing exact numerical reproduction. \\
HalluCitation Matters \citep{sakai2026hallucitation}; publication audit & 17,842 ACL-family papers; 741,656 references screened; candidates manually checked. & Paper-level count with at least one confirmed hallucinated reference. & Source/PDF/identifier checks; stop checking a paper after one confirmed error. & 295 papers (1.65\%) is a lower bound on affected papers, not a complete citation-level error count. \\
GhostCite \citep{xu2026ghostcite}; reference generation & 13 models, 40 CS areas, requested lists; 331,809 extracted references. & Validity proportion and JSON compliance reported separately. & Search/reasoning switches; manual checks of 400 valid and 400 invalid predictions. & The expected 375,440 entries is not the evaluated denominator; unparsed outputs are excluded from validity rates. \\
GhostCite \citep{xu2026ghostcite}; reference classification & 50 valid and 50 invalid entries; models judge bibliographic identity. & Accuracy and class-conditional detection. & Single-entry prompts with search/reasoning enabled. & The column named `Invalid Precision' divides by 50 actual invalid entries and is recall, not alarm precision. \\
GhostCite \citep{xu2026ghostcite}; archive audit & 56,381 papers; 2,199,409 references; retrieval candidates and PDFs. & Confirmed invalid references and affected-paper proportions. & 16 assistants; each candidate checked at least twice, with researcher review. & The original study's Section~VI-A reports 739 invalid citations, versus 738 in its Table~V. The authors state that the annual trend does not establish causality (their Section~VI-D). \\
GhostCite \citep{xu2026ghostcite}; researcher survey & 94 retained questionnaires about AI use and reference-checking habits. & Question-specific response proportions. & 97 responses, three inconsistent responses excluded; opt-in recruitment. & Self-reports do not measure experimentally observed invalid-reference insertion rates. \\
Zhang--Abernethy \citep{zhang2026detecting}; quotation errors & 250 natural statement--reference pairs; title, abstract, retrieved passage or PDF evidence. & Fully, partially or unsubstantiated accuracy; separate collapsed binary analysis. & Existing expert case labels; source-information and model controls. & Three-class accuracy 49.6--70.0\%; more information does not uniformly help. Relevant subclaims define each source's responsibility. \\
BibAgent \citep{li2026bibagent}; open-source miscitation & Citing statement and source full text; 6,350 expert-checked synthetic errors. & Acc-pass@3 requires validity plus accepted explanation; abstention counts as error. & Same-backbone full-text controls; taxonomy annotation and over 3,000 grader decisions checked separately. & All main-set entries are errors. Pass@3 is not single-attempt deployment accuracy or false-alarm measurement. \\
\bottomrule
\end{tabular}
\par\smallskip\noindent Table~\ref{tab:appendix-writing} (continued): Academic writing and scholarly communication protocols.\par
\endgroup
\end{minipage}
\end{strip}
\clearpage
\begin{strip}
\begin{minipage}{\textwidth}
\begingroup
\small
\setlength{\tabcolsep}{2.2pt}
\renewcommand{\arraystretch}{1.15}
\centering
\begin{tabular}{@{}>{\raggedright\arraybackslash}p{0.15\textwidth}>{\raggedright\arraybackslash}p{0.21\textwidth}>{\raggedright\arraybackslash}p{0.18\textwidth}>{\raggedright\arraybackslash}p{0.17\textwidth}>{\raggedright\arraybackslash}p{0.19\textwidth}@{}}
\toprule
\textbf{Work and task} & \textbf{Input/output evidence} & \textbf{Evaluator/metric} & \textbf{Calibration/controls} & \textbf{Findings/conditions} \\
\midrule
BibAgent \citep{li2026bibagent}; inaccessible sources & Downstream citing contexts substitute for unavailable source full text; 3,810-error subset. & Acc-pass@3, abstention and conditional token savings. & Same-backbone search control; witness-count/consistency thresholds. & Different evidence and subset from Open. Savings use jointly non-abstaining cases; indirect witnesses may share misconceptions (author limitation). \\
\multicolumn{5}{@{}l@{}}{\textbf{Posters, slides and scientific diagrams}} \\
Paper2Poster \citep{pang2025paper2poster}; poster quality & 100 paper--author-poster pairs; generated single-page posters. & AltCLIP, perplexity and six GPT-4o rubric dimensions. & Author-poster and generation baselines; five repeated evaluations of one system's 100 outputs. & AltCLIP is not direct aesthetic judgment. Single-system repeatability does not establish ranking stability across systems. \\
Paper2Poster \citep{pang2025paper2poster}; PaperQuiz & 100 questions per paper, split literal/interpretive; readers see only the poster. & Six VLM readers; raw QA and separately length-adjusted QA. & Full-paper/author-poster controls; one PhD participant, five papers, eight artifact conditions. & Length-adjusted score ranges to 200. Small human check limits the claimed general reader-proxy validation. \\
P2P/P2PEval \citep{sun2025p2p}; poster content & 121 paper--poster pairs; 1,738 author-poster-derived checklist items. & GPT-4o weighted checklist score, normalized to 0--100. & Three checklist annotators plus a fourth checker; visual items explicitly represented. & Content preservation and generic layout scores can change in opposite directions under reasoning-mode changes. \\
P2P/P2PEval \citep{sun2025p2p}; generic poster quality & Ten generic dimensions and human overall scores; generated posters. & Ten 0--5 dimensions; XGBoost calibration; separate pairwise preference. & 1,701 human scores; ten-fold validation reports $R^2=.92$. & Calibration predicts human totals; the original study's Section~3.2.1 does not specify whether folds are grouped by paper, poster, or individual rating. Strict wins and wins-or-ties differ. \\
SlideTailor \citep{zeng2025slidetailor}; preference-conditioned slides & Target paper, example paper--slides pair and PPTX template; editable ten-slide output. & Topic IoU, ordering, structure, preference and generic quality; six normalized scores. & 50 main cases; four people supply 60 ratings on 30 cases; cross-judge checks. & 40 wins/9 losses/11 ties: 81.63\% excludes ties. Potential 100,000 combinations are not generated test cases. \\
AutoFigure \citep{zhu2026autofigure}; scientific illustrations & Long scientific text and reference illustration; 300 test cases within 3,300 pairs. & Eight VLM submetrics; automatic blind comparisons; human ratings/rankings. & Ten first authors evaluate 21 papers; separate 95-output human--judge calibration. & The original study's Table 2 reports automatic evaluation; human evaluation is reported separately. Authors document residual text errors after correction; human adoption denominator is unclear. \\
PaperBanana \citep{zhu2026paperbanana}; method diagrams & Method text/caption and reference library; 292 reference and 292 test cases. & Four reference-comparison dimensions; hierarchical overall win/tie/loss score. & 50 judge-calibration cases; separate 50-case three-person blind preference study. & Reference baseline 50 is defined by scoring. Overall 60.2 coexists with faithfulness 45.8; authors acknowledge this shortfall. \\
\bottomrule
\end{tabular}
\par\smallskip\noindent Table~\ref{tab:appendix-writing} (continued): Academic writing and scholarly communication protocols.\par
\endgroup
\end{minipage}
\end{strip}
\clearpage
\begin{strip}
\begin{minipage}{\textwidth}
\begingroup
\small
\setlength{\tabcolsep}{2.2pt}
\renewcommand{\arraystretch}{1.15}
\centering
\begin{tabular}{@{}>{\raggedright\arraybackslash}p{0.15\textwidth}>{\raggedright\arraybackslash}p{0.21\textwidth}>{\raggedright\arraybackslash}p{0.18\textwidth}>{\raggedright\arraybackslash}p{0.17\textwidth}>{\raggedright\arraybackslash}p{0.19\textwidth}@{}}
\toprule
\textbf{Work and task} & \textbf{Input/output evidence} & \textbf{Evaluator/metric} & \textbf{Calibration/controls} & \textbf{Findings/conditions} \\
\midrule
PaperBanana \citep{zhu2026paperbanana}; statistical charts & Underlying table data; Matplotlib-generated charts; ChartMimic-derived cases. & Reported chart-quality comparison under the chart protocol. & 240 reference and 240 test charts across seven types; baseline comparison. & Separate code-rendered task, not the method-diagram test; reported overall improvement is 4.1 points. \\
SciFig \citep{huang2026scifig}; editable method diagrams & Method/caption to draw.io XML, including embedded PNG components; 435 tests. & 50-question VQA, visual/OCR similarity, 20-item rubric and eight design heuristics. & 1,784 rubric-development cases; 60 people, 10,000 blind comparisons; human alignment checks. & Supports component-level editing through XML with embedded PNG assets; human comparisons assess preferences for generated diagrams. \\
SridBench \citep{chang2025sridbench}; scientific illustrations & 1,120 image--caption--section triples across 13 subject directions. & Six GPT-4o 1--5 dimensions relative to original diagrams. & 100 cases compare two model judges with human ratings; two generators fully evaluated. & Authors report textual omissions and scientific errors. Low means do not define an absolute zero-capability threshold. \\
\bottomrule
\end{tabular}
\par\smallskip\noindent Table~\ref{tab:appendix-writing} (continued): Academic writing and scholarly communication protocols.\par
\endgroup
\end{minipage}
\end{strip}

\clearpage
\subsection{Automatic Peer Review Protocols}
The records distinguish review-content evaluators, manuscript-grounded criticism, multimodal checking, and feedback use or aggregation. Natural weaknesses, injected errors and decision labels supply different references; a paper may therefore contribute several task-specific rows.

\begin{strip}
\begin{minipage}{\textwidth}
\begingroup
\small
\setlength{\tabcolsep}{2.2pt}
\renewcommand{\arraystretch}{1.15}
\centering
\begin{tabular}{@{}>{\raggedright\arraybackslash}p{0.15\textwidth}>{\raggedright\arraybackslash}p{0.21\textwidth}>{\raggedright\arraybackslash}p{0.18\textwidth}>{\raggedright\arraybackslash}p{0.17\textwidth}>{\raggedright\arraybackslash}p{0.19\textwidth}@{}}
\toprule
\textbf{Work and task} & \textbf{Input/output evidence} & \textbf{Evaluator/metric} & \textbf{Calibration/controls} & \textbf{Findings/conditions} \\
\midrule
\multicolumn{5}{@{}l@{}}{\textbf{Review content, evaluators and calibration}} \\
ReviewEval \citep{garg2025revieweval}; review dimensions & Manuscript, official reviews and venue rules; generated reviews and automatic rebuttals. & Semantic/topic alignment; question-level support; depth, constructiveness and guideline scores. & 120-paper corpus; five experts compare outputs on a 30-paper subset across four preference dimensions. & Preference calibration does not independently label factual support decisions. ReviewAgent is a separate evaluated generator. \\
GEM/GRE-bench \citep{xu2025benchmarkinga}; shared information & Candidate and peer-reference reviews; optional abstract/author-strengths context. & Probability-model estimate of shared information; GEM-S conditions out supplied background. & Teacher correlations on about 180 proposal reviews; separate degradation/manipulation tests on 300 ICLR papers. & Requires peer references, not unique gold. GEM-S teacher correlation .479 is below LMExaminer's .537; robustness tests measure another property. \\
Beyond Rating \citep{li2026rating}; multidimensional review alignment & 1,000 papers, 3,666 high-confidence/low-disagreement reviews; extracted atomic points. & Summary coverage, point precision/max-recall, focus KL, QuestionScore, Binoculars and rating MAE. & Alternative extraction-model check; shared prompts for generic models. & Binoculars lacks independent depth calibration; QuestionScore checks retained text after removing some sections. \\
MindA \citep{shin2025minda}; focus alignment & Review/meta-review points coded by manuscript target and evaluation aspect. & Group-distribution KL; per-paper target--aspect F1; text-similarity controls. & 327 points from 68 papers compared with author-discussion labels; near-submission versions for accepted-paper supplement. & Focus means category frequency, not attention weights or critique truth. Abstract/main-text sample counts differ. \\
Liang et al. \citep{liang2023can}; feedback overlap & Nature/ICLR papers and reviews; GPT-4 comments from the first 6,500 input tokens. & Extracted-point overlap, hit rate and matching coefficients. & Human extraction/matching validation; shuffled-paper specificity control; matched point-count comparison. & Reported overlap measures shared feedback concerns. Rejected-paper overlap and total review counts have version-internal conflicts. \\
Liang et al. \citep{liang2023can}; perceived usefulness & Uploaded manuscripts, model feedback and questionnaires from 308 researchers. & Perceived helpfulness and comparisons with prior human feedback. & Prospective opt-in users from AI/computational biology; user survey separate from corpus matching. & 57.4\% rate feedback helpful/very helpful. The 82.4\% comparison is with at least some human feedback, not the average expert. \\
\bottomrule
\end{tabular}
\captionof{table}{Automatic peer-review evaluation protocols.}\label{tab:appendix-peer-review}
\endgroup
\end{minipage}
\end{strip}
\clearpage
\begin{strip}
\begin{minipage}{\textwidth}
\begingroup
\small
\setlength{\tabcolsep}{2.2pt}
\renewcommand{\arraystretch}{1.15}
\centering
\begin{tabular}{@{}>{\raggedright\arraybackslash}p{0.15\textwidth}>{\raggedright\arraybackslash}p{0.21\textwidth}>{\raggedright\arraybackslash}p{0.18\textwidth}>{\raggedright\arraybackslash}p{0.17\textwidth}>{\raggedright\arraybackslash}p{0.19\textwidth}@{}}
\toprule
\textbf{Work and task} & \textbf{Input/output evidence} & \textbf{Evaluator/metric} & \textbf{Calibration/controls} & \textbf{Findings/conditions} \\
\midrule
Zhou et al. \citep{zhou2024llm}; rating inference & PeerRead paper text or human review text; aspect/overall reference scores. & Score error and correlation. & Paper-input versus review-input conditions; zero-/few-shot prompts. & Rating recovery from reviews and assessment from manuscripts yield different correlations. \\
Zhou et al. \citep{zhou2024llm}; review generation & ASAP manuscripts to aspect/sentiment-annotated reviews; human reference reviews. & ROUGE, BERTScore, BLANC; model and human relevance/informativeness ratings. & 300-paper GPT-3.5 sample; 50-paper GPT-4 human analysis. & BERTScore has near-zero reported correlation with human relevance/informativeness in this sample. \\
Zhou et al. \citep{zhou2024llm}; RR-MCQ & 196 multiple-answer questions from 14 papers' review--rebuttal discussions. & Per-option accuracy and whole-question exact correctness. & Shuffled options; section-extraction variants; disputed source discussions excluded. & GPT-4 pipeline: .710 per option, .276 per question. Reported F1 conflicts with the listed precision and recall. \\
\multicolumn{5}{@{}l@{}}{\textbf{Manuscript-grounded criticism and error finding}} \\
CLAIMCHECK \citep{ou2025claimchecka}; claim association & Manuscript plus human weakness; locate up to three directly criticized claim spans. & Edit/exact-span F1. & Expert links; pilot agreement separate from mainly single-annotator main set. & 41 papers, 168 weaknesses; 120 weaknesses linked to 154 target claims. Pilot human scores are not same-test upper bounds. \\
CLAIMCHECK \citep{ou2025claimchecka}; weakness labeling/editing & Known weakness and target claims; classify and rewrite the criticism. & Krippendorff's alpha for labels; rewriting preference and meaning change. & 20 rewrites per model; two judges each, randomized order. & Supplying the weakness distinguishes this task from independent discovery; preferred rewrites may change meaning. \\
CLAIMCHECK \citep{ou2025claimchecka}; claim verification & Manuscript and claim, without supplied weakness; generate/check criticism. & Label agreement; Same, Similar or Different reference-weakness matching. & Compared with weakness-given condition using the same task family. & Low reference agreement is reported; a different criticism is not automatically an incorrect one. \\
Peerispect \citep{ghorbanpour2026peerispect}; CMC & 500 extracted manuscript claims from 50 ICLR papers; all supported by construction. & Supported-label recovery accuracy; evidence retrieval recall. & Retriever/reranker and 3B/7B verifier comparisons; top-three passages. & Single-positive-class accuracy does not test support-versus-nonsupport discrimination; retrieval is separately measured. \\
Peerispect \citep{ghorbanpour2026peerispect}; RRC & 150 real review claims from 25 papers; human labels informed by manuscript and rebuttal. & Four-class verification accuracy. & Same modular comparisons; no unique oracle paragraph, hence no RRC retrieval recall. & Best reported four-class accuracy is .287 on this real-review set. \\
FLAWS \citep{xi2025flaws}; injected-error localization & 713 paper--error pairs; PDF, injected claim-invalidating edits and associated reference spans. & Top-k localization by span similarity or semantic match; bootstrap intervals. & 58-error human audit; 253 agreed detection judgments calibrate scoring; insertion-source regression controls. & 57/58 sampled errors valid; scorer precision .91/recall .98 on agreed cases. k is ranked candidate depth, not reruns. \\
SPOT \citep{son2025when}; acknowledged-error discovery & 83 historical papers with 91 author-acknowledged, self-contained errors; full text/images. & Semantic reference matching; precision, recall and pass@K. & Eight runs per paper/model; 48-paper text/image comparison; domain-expert case audits. & Reference-outside true error documented. Unmatched findings count as false positives, so matching precision is not critique truth rate. \\
\bottomrule
\end{tabular}
\par\smallskip\noindent Table~\ref{tab:appendix-peer-review} (continued): Automatic peer-review evaluation protocols.\par
\endgroup
\end{minipage}
\end{strip}
\clearpage
\begin{strip}
\begin{minipage}{\textwidth}
\begingroup
\small
\setlength{\tabcolsep}{2.2pt}
\renewcommand{\arraystretch}{1.15}
\centering
\begin{tabular}{@{}>{\raggedright\arraybackslash}p{0.15\textwidth}>{\raggedright\arraybackslash}p{0.21\textwidth}>{\raggedright\arraybackslash}p{0.18\textwidth}>{\raggedright\arraybackslash}p{0.17\textwidth}>{\raggedright\arraybackslash}p{0.19\textwidth}@{}}
\toprule
\textbf{Work and task} & \textbf{Input/output evidence} & \textbf{Evaluator/metric} & \textbf{Calibration/controls} & \textbf{Findings/conditions} \\
\midrule
LimitGen \citep{xu2025can}; synthetic limitations & 500 NLP papers, 1,000 manually checked perturbation cases; given target aspect. & Top-three limitation identification; match-gated specificity/relevance scores. & Human correction of 112 cases; sampled human checks; with/without retrieval. & Perturbation cases are not 1,000 independent papers; authors limit the scope of cross-domain extension. \\
LimitGen \citep{xu2025can}; natural limitations & 1,000 ICLR papers and extracted review weaknesses; aspect-conditioned output. & Reference precision/recall/Jaccard and match-gated quality; independent human dimensions. & Human faithfulness, soundness and importance ratings initially hide references; retrieval comparison. & Unmatched plausible limitations receive automatic zero. Human quality evidence is separate from reference coverage. \\
ReviewScore \citep{ryu2025reviewscore}; claims and questions & Review points plus manuscript text/tables; expert factuality and answerability labels. & Misinformed-detection F1 and five-point weighted kappa. & Three-person annotation; rebuttal-input comparison. & Main set has 573 points from 40 papers; the 657-point total includes pilot annotations. Questions check whether the paper already answers them. \\
ReviewScore \citep{ryu2025reviewscore}; argument premises & Arguments reconstructed into premises/conclusion, formal logic and natural language. & SAT validity; premise factuality; conjunction or weighted aggregation. & Independent human reconstruction faithfulness; low-faithfulness reconstructions gated out. & Authors report aggregation can propagate errors or reduce agreement; logical validity and premise truth are separate checks. \\
Counterfactual review \citep{dycke2026automatic}; reasoning-sensitive response & 133 papers, 391 critical and 540 neutral edits; generated reviews and ratings. & Average treatment effects on review aspects, positive assertions and scores; mixed-effects tests. & Neutral rewriting control; oracle positive control; 92 edits from ten papers human-checked. & No significant extra critical-edit response in tested statistics. Full-pipeline response does not isolate reasoning from visibility/localization. \\
\multicolumn{5}{@{}l@{}}{\textbf{Multimodal manuscript evidence}} \\
PRISMM-Bench \citep{selch2026prismmbench}; identification & 384 verified text/figure/table inconsistencies from 353 manuscripts; four-choice questions. & Accuracy; visual reliance relative to no-context performance. & Focused, page and document inputs; natural-language versus JSON options; matched human subset. & Focused evidence is already localized. Structured options reduce, but do not eliminate, answer-wording shortcuts. \\
PRISMM-Bench \citep{selch2026prismmbench}; remedy selection & 384 inconsistencies; Target--Action remedy options and manuscript evidence. & Four-choice accuracy under three context levels. & Target--Action JSON options converted from manually refined identification choices. & Tests choosing a repair, not executing one; authors explicitly define this endpoint. \\
PRISMM-Bench \citep{selch2026prismmbench}; visual matching & 192 visual--visual inconsistency pairs; focused crops and candidate matches. & Four-choice matching accuracy. & Focused context only; underlying evidence manually verified. & The total of 960 task instances includes repeated source inconsistencies across the three protocols. \\
MMReview \citep{gao2025mmreview}; summary (S) & 240 manuscripts; generated summaries versus human-review summaries. & BARTScore and LLM semantic/content comparison. & Text, text+figures and page-image conditions. & Task asks for a summary in the model's own words, excluding subjective assessment. \\
\bottomrule
\end{tabular}
\par\smallskip\noindent Table~\ref{tab:appendix-peer-review} (continued): Automatic peer-review evaluation protocols.\par
\endgroup
\end{minipage}
\end{strip}
\clearpage
\begin{strip}
\begin{minipage}{\textwidth}
\begingroup
\small
\setlength{\tabcolsep}{2.2pt}
\renewcommand{\arraystretch}{1.15}
\centering
\begin{tabular}{@{}>{\raggedright\arraybackslash}p{0.15\textwidth}>{\raggedright\arraybackslash}p{0.21\textwidth}>{\raggedright\arraybackslash}p{0.18\textwidth}>{\raggedright\arraybackslash}p{0.17\textwidth}>{\raggedright\arraybackslash}p{0.19\textwidth}@{}}
\toprule
\textbf{Work and task} & \textbf{Input/output evidence} & \textbf{Evaluator/metric} & \textbf{Calibration/controls} & \textbf{Findings/conditions} \\
\midrule
MMReview \citep{gao2025mmreview}; strengths (SE) & 238 manuscripts; generated acceptance-oriented strengths versus human points. & BARTScore and LLM reference comparison. & Shared modality controls; manually verified reference extraction. & Quality, clarity, significance and originality guide the output. \\
MMReview \citep{gao2025mmreview}; weaknesses (WE) & 240 manuscripts; generated rejection-oriented criticisms versus human points. & BARTScore and LLM reference comparison. & Same modality comparisons as summary/strength tasks. & Scores concern coverage and semantic similarity to selected human criticisms. \\
MMReview \citep{gao2025mmreview}; soundness (SS) & 32 manuscripts with human soundness scores; model assesses technical/evidential support. & MAE on integer 1--4 scores. & Modality and model comparisons. & This task's 32 cases are a subset, not the full 240-paper corpus. \\
MMReview \citep{gao2025mmreview}; presentation (PS) & 32 manuscripts with human presentation ratings; clarity and organization. & MAE on integer 1--4 scores. & Modality and model comparisons. & Presentation and technical soundness have separate reference ratings. \\
MMReview \citep{gao2025mmreview}; conditional decision (CD) & 176 cases with human-written reviews supplied; infer overall paper score. & MAE on 1--10 ratings. & Compared with paper-input decision tasks, retaining the input difference. & Extra human reviews supply evaluative information absent from DD/CoD. \\
MMReview \citep{gao2025mmreview}; direct decision (DD) & 176 manuscripts; directly produce overall score. & MAE on 1--10 ratings. & Direct-input baseline for structured CoD. & Direct paper assessment differs from review-conditioned score recovery. \\
MMReview \citep{gao2025mmreview}; structured decision (CoD) & 176 manuscripts; summary, strengths/weaknesses, subscores, then overall score. & Overall-score MAE. & Same task family as DD; structured intermediate reasoning. & Improvement over DD varies by model; Tiny/Middle group means are worse in the reported table. \\
MMReview \citep{gao2025mmreview}; meta decision (MD) & 240 cases with multiple human reviews; binary accept/reject output. & Decision accuracy against recorded outcomes. & Guided review synthesis rather than only averaging numerical scores. & Paper-alone decisions and human-review aggregation use different evidence. \\
MMReview \citep{gao2025mmreview}; pairwise rank (PR) & 96 paper pairs from different acceptance tiers; choose relative preference. & Accuracy against recorded tier ordering. & Each pair repeated with reversed input order. & Preference target is conference-tier alignment. \\
MMReview \citep{gao2025mmreview}; fake strengths (FS) & 240 cases containing polarity-reversed human-review claims; manuscript evidence. & MAE as specified in the original study's Table 2 for numerical evaluations. & GPT-4o antonymic rewrites with manual semantic checks. & Constructed false-positive assessments, separate from open-ended strength generation. \\
MMReview \citep{gao2025mmreview}; fake weaknesses (FW) & 238 cases with polarity-reversed human-review claims; manuscript evidence. & MAE as specified in the original study's Table 2. & Same construction and manual verification family as FS. & Constructed criticism assessment, not an unrestricted error-discovery task. \\
MMReview \citep{gao2025mmreview}; prompt injection (PI) & 176 manuscripts with hidden instructions; otherwise the CoD setup. & Score deviation, increase/decrease frequency and conditional change magnitude. & Clean-input control; PDF-as-image excluded from this text-injection task. & Change among increasing cases is not average change over all cases. \\
\bottomrule
\end{tabular}
\par\smallskip\noindent Table~\ref{tab:appendix-peer-review} (continued): Automatic peer-review evaluation protocols.\par
\endgroup
\end{minipage}
\end{strip}
\clearpage
\begin{strip}
\begin{minipage}{\textwidth}
\begingroup
\small
\setlength{\tabcolsep}{2.2pt}
\renewcommand{\arraystretch}{1.15}
\centering
\begin{tabular}{@{}>{\raggedright\arraybackslash}p{0.15\textwidth}>{\raggedright\arraybackslash}p{0.21\textwidth}>{\raggedright\arraybackslash}p{0.18\textwidth}>{\raggedright\arraybackslash}p{0.17\textwidth}>{\raggedright\arraybackslash}p{0.19\textwidth}@{}}
\toprule
\textbf{Work and task} & \textbf{Input/output evidence} & \textbf{Evaluator/metric} & \textbf{Calibration/controls} & \textbf{Findings/conditions} \\
\midrule
\multicolumn{5}{@{}l@{}}{\textbf{Usefulness, intervention and opinion aggregation}} \\
RevUtil \citep{sadallah2025good}; author-oriented review utility & Review point only, without manuscript; 1,430 human-labeled points and synthetic training labels. & Four 1--5 aspects; weighted kappa, correlation; claim-detection F1. & Three annotators per point; full human set and majority-agreement subsets reported separately. & Grounding is specificity of reference to the paper; verifiability is provision of reasons. Neither checks those reasons against the manuscript. \\
Thakkar et al. \citep{thakkar2025can}; randomized feedback & ICLR review versions, random assignment, delivered feedback and subsequent discussion. & Edits, length, uptake, discussion and acceptance statistics. & 22,467 selected/22,364 control; intention-to-treat separate from received-feedback analysis. & 18,946 actually received feedback after gating. Assignment and receipt have different denominators. \\
Thakkar et al. \citep{thakkar2025can}; revised-review preference & 100 before/after review pairs with 3--4 feedback items and uptake above .60. & Blind researcher preference for revised versus initial review. & Two AI researchers; selected high-uptake revisions. & 89\% preference applies to this selected subset, not every assigned or delivered review. \\
ORSUM \citep{zeng2023scientific}; meta-review generation & Title, abstract, review set and given decision; generated consensus/dispute synthesis. & Text/factuality metrics; model rubrics; human binary informativeness, soundness, consistency and faithfulness. & Three annotators; 50 difficult borderline cases; three automatic runs. & Reviews capped at 300 tokens; no rebuttal. Decision is an input; human/LexRank perfect faithfulness is assumed, not measured. \\
\bottomrule
\end{tabular}
\par\smallskip\noindent Table~\ref{tab:appendix-peer-review} (continued): Automatic peer-review evaluation protocols.\par
\endgroup
\end{minipage}
\end{strip}

\clearpage
\subsection{End-to-End Research Protocols}
System demonstrations and review-feedback studies are followed by multi-task infrastructure, connected research subprotocols, and rediscovery. The suite, stage and full-run rows preserve their different denominators.

\begin{strip}
\begin{minipage}{\textwidth}
\begingroup
\small
\setlength{\tabcolsep}{2.2pt}
\renewcommand{\arraystretch}{1.15}
\centering
\begin{tabular}{@{}>{\raggedright\arraybackslash}p{0.15\textwidth}>{\raggedright\arraybackslash}p{0.21\textwidth}>{\raggedright\arraybackslash}p{0.18\textwidth}>{\raggedright\arraybackslash}p{0.17\textwidth}>{\raggedright\arraybackslash}p{0.19\textwidth}@{}}
\toprule
\textbf{Work and task} & \textbf{Input/output evidence} & \textbf{Evaluator/metric} & \textbf{Calibration/controls} & \textbf{Findings/conditions} \\
\midrule
\multicolumn{5}{@{}l@{}}{\textbf{System demonstrations and feedback}} \\
AI Scientist \citep{lu2024aiscientist}; connected research & Baseline code, plot scripts and LaTeX templates; ideas, code changes, experiments, plots and papers. & Task-native losses/KL; funnel counts; automated review scores and costs. & Three lightweight ML domains; multiple backbones; author case inspection. & Experiment completion and compiled-paper counts differ. Reported threshold crossings are automated-review outcomes. \\
AI Scientist \citep{lu2024aiscientist}; internal reviewer calibration & 500 ICLR 2022 papers and decisions; text-only model reviews. & Decision accuracy, balanced accuracy, F1/AUC and rating correlation. & Reflection, ensemble and threshold variants; published NeurIPS human statistics as a reference. & Cross-conference human comparison and model-to-mean versus person-to-person correlations do not establish same-condition human-level performance. \\
DeepScientist \citep{weng2025deepscientist}; sustained research & Reproduced baselines, findings memory, implementations and task results across three AI tasks. & Accuracy, throughput, AUROC/latency; idea-to-validation-to-progress counts. & Human result verification; random-idea filtering comparison; failed-implementation audit. & About 5,000 ideas, 1,100 validations, 21 progress findings; 20,000+ GPU hours. Historical publication spans are not measured human research time. \\
DeepScientist \citep{weng2025deepscientist}; generated papers & Five system papers and external generated papers; automatic and expert reviews. & DeepReviewer ratings; three-person simulated review panel and agreement. & Three independent human ratings per paper; no rebuttal. & Automatic mean rating 5.90 and human mean 5.00 are different channels. Simulated acceptance is not conference acceptance. \\
CycleResearcher \citep{weng2025cycleresearcher}; simulated research feedback & Bibliography/abstracts to long research papers with simulated experiments; review-guided refinement. & Automated paper rubrics and simulated acceptance; selected expert review. & Independent reward-model check; three NLP experts compare selected outputs anonymously. & Main protocol explicitly simulates results. The original study's Appendix I separately demonstrates one executed study; it is not the main evaluation. \\
CycleReviewer \citep{weng2025cycleresearcher}; review calibration & Review-5k paper/review data; generated reviews and predicted decisions. & Proxy MAE/MSE against other reviewers' means; decision accuracy/macro-F1. & Human-reviewer proxy baseline under n versus n-1 aggregation. & Agreement with aggregate ratings is the proxy target. Reported percentage reductions conflict with rounded table values. \\
\bottomrule
\end{tabular}
\captionof{table}{End-to-end scientific research system evaluation protocols.}\label{tab:appendix-end-to-end}
\endgroup
\end{minipage}
\end{strip}
\clearpage
\begin{strip}
\begin{minipage}{\textwidth}
\begingroup
\small
\setlength{\tabcolsep}{2.2pt}
\renewcommand{\arraystretch}{1.15}
\centering
\begin{tabular}{@{}>{\raggedright\arraybackslash}p{0.15\textwidth}>{\raggedright\arraybackslash}p{0.21\textwidth}>{\raggedright\arraybackslash}p{0.18\textwidth}>{\raggedright\arraybackslash}p{0.17\textwidth}>{\raggedright\arraybackslash}p{0.19\textwidth}@{}}
\toprule
\textbf{Work and task} & \textbf{Input/output evidence} & \textbf{Evaluator/metric} & \textbf{Calibration/controls} & \textbf{Findings/conditions} \\
\midrule
\multicolumn{5}{@{}l@{}}{\textbf{Suites, research stages and task-native outcomes}} \\
AstaBench \citep{bragg2026astabench}; multi-task suite & 11 adapted benchmarks, 2,400+ problems; literature, code, analysis and discovery tasks. & Task-specific program/model scorers; macro aggregates and token-cost metadata. & Standard tools/environments; custom-agent tooling disclosed by category; frozen price table. & 57 agent configurations are not a complete 57-by-11 matrix. Suite macro score is not a connected research-run success rate. \\
AstaBench E2E-Bench \citep{bragg2026astabench}; research steps & Expert-revised detailed research questions; reports, code and generated artifacts. & Rubric step passes require no conflicting evidence and at least one supporting source; all-step completion separately. & 40 test/10 validation tasks; Faker negative control; human grader check of 50 development-set rubric items. & Human check: 92\% grading correctness. The original study's Table~21 separately reports 16\% report-apparent passes contradicted by code/artifacts. Faker's whole-task completion is zero. \\
AstaBench E2E-Hard \citep{bragg2026astabench}; harder research tasks & HypER-generated/refined questions without the same simplification; three artifact channels. & Mean step score and all-required-step completion; per-task cost. & Separate 40 test/10 validation tasks; same cross-artifact rubric logic and Faker control. & Panda step score 68.2 and whole-task completion .03 have different units; not the full-suite result. \\
MLR-Bench \citep{chen2025mlrbench}; ideas and proposals & 201 workshop-derived research tasks; ideas and proposals, with shared sampled upstream input in stage tests. & Phase-specific MLR-Judge rubrics; two model scores averaged. & Six generation models; retrieval between idea and proposal phases. & 201 is the early-stage task count, not 201 completed end-to-end studies. \\
MLR-Bench \citep{chen2025mlrbench}; experiments and writing & Ten selected implementable tasks; proposals, code, logs, figures, reports and papers. & Experiment soundness/completeness and paper rubrics. & Common stage inputs; coding/writing model comparisons; four RTX 3090 environment. & Authors find unverified synthetic/placeholder-data results in eight of ten Claude Code tasks; limited to that configuration. \\
MLR-Bench \citep{chen2025mlrbench}; end-to-end and judge checks & Ten-task connected runs; final papers plus execution evidence; expert ratings. & Two-judge overall scores; human--model versus human--human absolute differences. & Ten experienced experts, two ratings per paper; experts mainly read papers. & Non-significant discrepancy-distribution tests do not establish equivalence. Human/model evidence permissions differ. \\
ResearchGym \citep{garikaparthi2026researchgym}; research execution & Five repositories, 39 subtasks; source method removed, baselines/data/graders retained. & Task-native scores, agent/SOTA normalization, completion and improvement rates; means/best@3. & Three initial runs/task, 12 h, A100 80GB, \$10 API; source-method reinsertion check; post-run integrity audit. & Main agent improves visible baseline in 1/15 runs; 26.5\% mean subtask completion. Best-run extensions have additional budget. \\
AIRS-Bench \citep{lupidi2026airsbench}; solution construction & 20 problem/dataset/metric tasks, no initial solution code; prediction CSV on hidden test labels. & Native scores; valid submissions; normalized score and Elo; failures score zero. & At least ten runs per configuration and task; 24 h and one H200; human checks of task configuration. & Four tasks have at least one result above literature SOTA. Valid-submission rates conflict in text/figure; training exposure is not ruled out. \\
\bottomrule
\end{tabular}
\par\smallskip\noindent Table~\ref{tab:appendix-end-to-end} (continued): End-to-end scientific research system evaluation protocols.\par
\endgroup
\end{minipage}
\end{strip}
\clearpage
\begin{strip}
\begin{minipage}{\textwidth}
\begingroup
\small
\setlength{\tabcolsep}{2.2pt}
\renewcommand{\arraystretch}{1.15}
\centering
\begin{tabular}{@{}>{\raggedright\arraybackslash}p{0.15\textwidth}>{\raggedright\arraybackslash}p{0.21\textwidth}>{\raggedright\arraybackslash}p{0.18\textwidth}>{\raggedright\arraybackslash}p{0.17\textwidth}>{\raggedright\arraybackslash}p{0.19\textwidth}@{}}
\toprule
\textbf{Work and task} & \textbf{Input/output evidence} & \textbf{Evaluator/metric} & \textbf{Calibration/controls} & \textbf{Findings/conditions} \\
\midrule
InnovatorBench \citep{wu2025innovatorbench}; long-horizon LLM research & 20 tasks/six categories; workspaces with core innovation removed; data, models or code submitted. & External task-specific metrics; normalized final/best scores, weighted aggregates and cost. & 13 task builders reproduce references; common ReAct; at most three intermediate evaluations; hints disabled in main runs. & Reference solution is near 80, not 100. Best means best submission within a run. Per-task compute/time and network permissions vary. \\
\multicolumn{5}{@{}l@{}}{\textbf{Scientific rediscovery and diagnosis}} \\
FIRE-Bench \citep{wang2026firebench}; core rediscovery & High-level question/data/conditions; original implementation and findings hidden; generated conclusions versus reference findings. & Atomic-claim entailment precision/recall/F1, averaged over runs. & 30 core tasks, three runs/system; human claim-extraction check; inspected trajectories for source access. & Best mean F1 46.7 is a claim score, not task success percentage. Added human calibration covers extraction, not the final entailment scorer. \\
FIRE-Bench \citep{wang2026firebench}; cross-domain extension & Ten additional tasks from the 100-paper source pool; generated findings and reference claims. & Same claim-level scoring, separate cross-domain results. & Three systems; Claude Code excluded for budget reasons; repeated runs. & Other 60 source papers are parsed/released but not run in this study. All three reported means are below 25 F1. \\
FIRE-Bench \citep{wang2026firebench}; failure attribution & Execution trajectories, code, conclusions and original paper; unmatched/incorrect findings. & Four stages, 16 error classes; false-positive categories including alternative conclusions. & LLM preliminary annotation followed by author item-wise review. & Authors identify planning/conclusion failures and occasional plausible alternative conclusions; reference-outside conclusions can be penalized. \\
\bottomrule
\end{tabular}
\par\smallskip\noindent Table~\ref{tab:appendix-end-to-end} (continued): End-to-end scientific research system evaluation protocols.\par
\endgroup
\end{minipage}
\end{strip}

\end{document}